\documentclass{article}

\PassOptionsToPackage{numbers,sort}{natbib}

\usepackage[final]{neurips_2022}

\usepackage[utf8]{inputenc} 
\usepackage[T1]{fontenc}    
\usepackage{hyperref}       
\usepackage{url}            
\usepackage{booktabs}       
\usepackage{amsfonts}       
\usepackage{nicefrac}       
\usepackage{microtype}      
\usepackage{xcolor}         

\usepackage{amsmath}
\usepackage{amssymb}
\usepackage{amsthm}
\usepackage{graphicx}
\usepackage{enumerate}
\usepackage{subfig}
\usepackage{pgfplots}
\pgfplotsset{width=10cm,compat=1.9}
\usepgfplotslibrary{groupplots}

\newtheorem{theorem}{Theorem}
\newtheorem{lemma}{Lemma}
\newtheorem{definition}{Definition}

\title{On Image Segmentation With Noisy Labels: \\ Characterization and Volume Properties of the Optimal Solutions to Accuracy and Dice}

\author{
  Marcus Nordstr\"om \thanks{Author is also affiliated with RaySearch Laboratories.}\\
  Department of Mathematics\\
  KTH Royal Institute of Technology\\
  Stockholm, Sweden \\
  \texttt{marcno@kth.se} \\
  \And
  Henrik Hult \\
  Department of Mathematics\\
  KTH Royal Institute of Technology\\
  Stockholm, Sweden \\
  \texttt{hult@kth.se} \\
  \And
  Jonas S\"oderberg \\
  Department of Machine Learning \\
  RaySearch Laboratories \\
  Stockholm, Sweden \\
  \texttt{jonas.soderberg@raysearchlabs.com} \\
  \And
  Fredrik L\"ofman \\
  Department of Machine Learning \\
  RaySearch Laboratories \\
  Stockholm, Sweden \\
  \texttt{fredrik.lofman@raysearchlabs.com} \\
}

\begin{document}

\maketitle

\begin{abstract}
We study two of the most popular performance metrics in medical image segmentation, Accuracy and Dice, when the target labels are noisy. For both metrics, several statements related to characterization and volume properties of the set of optimal segmentations are proved, and associated experiments are provided. Our main insights are: (i) the volume of the solutions to both metrics may deviate significantly from the expected volume of the target, (ii) the volume of a solution to Accuracy is always less than or equal to the volume of a solution to Dice and (iii) the optimal solutions to both of these metrics coincide when the set of feasible segmentations is constrained to the set of segmentations with the volume equal to the expected volume of the target.
\end{abstract}

\section{Introduction}
One of the most central problems in medical image analysis is to identify the region of an image associated with a certain target structure. This problem, referred to as image segmentation or delineation, is often very time consuming to solve manually.
Consequently, there is great interest in the development of methods that can assist in automation of the procedure. Since 2015, it has become increasingly popular to address the segmentation problem using machine learning based methods, and in particular, fully convolutions neural networks with U-net architecture.
Such methods commonly dominate the winning submissions to segmentation contests and are backed by a large base of supporting literature~\cite{ronneberger2015u,cciccek20163d,drozdzal2016importance,jiang2019two,litjens2017survey,sudre2017generalised}.

Despite the success, performance of these methods will, like any machine learning method, depend on the quality of the available data~\cite{yu2020robustness, chi2020deep}. Since it is well known that the data commonly used in practice is produced by medical practitioners that delineate structures in an inconsistent manner~\cite{nyholm2018mr}, it is important to understand the impact of label noise.
One way to study the influence of label noise is to consider how the noise impacts segmentations that are theoretically optimal with respect to the metric used for measuring performance. 
Even if theoretically optimal solutions may not be attainable in practice, studying them gives important insights into what the effect label noise has on the particular metric and any training method that is designed to maximize it.

Arguably, the most simple and classical choice of metric is Accuracy; the fraction of the image that is correctly delineated.
This metric is most commonly targeted by taking a $1/2$-threshold of predictions from a model trained with the cross-entropy loss~\cite{nordstrom2020calibrated}.
However, since Accuracy may not reflect the desired behaviour when the data is unbalanced, that is, when the target structure is much smaller than the background, alternative metrics are often preferred.
The most popular such alternative is the S{\o}rensen-Dice coefficient, or Dice for short and is related to the $\mathrm{F}_1$-metric used in binary classification.
This metric is most commonly targeted by taking a $1/2$-threshold of predictions from a model trained with the soft-Dice loss, a smoothed version of the Dice metric~\cite{nordstrom2020calibrated}.
Other examples considered in the literature include the Jaccard index and variations of the Haussdorff metric~\cite{taha2015metrics}.


In this work, we conduct a theoretical investigation of the effect label noise has on optimal segmentations with respect to the performance metrics Accuracy and Dice.
Because the volume of a proposed segmentation may be used for important properties such as estimating the size of a tumor~\cite{dubben1998tumor}, we pay special attention to the effect noise may have on the volume of the optimal segmentations.

\paragraph{Contributions:}
A characterization of all optimal solutions to Accuracy and Dice when the target is noisy is provided.
This characterization is used to analyze the volume of the optimal solutions and we prove:
(i) sharp upper and lower bounds on the volume of optimal segmentations with respect to Accuracy and Dice, 
(ii) that the volume of the optimal solutions to Accuracy always is less or equal to the volume of the optimal solutions to Dice and (iii) that the optimal solutions to both metrics coincide when the volume is held fix.
We also show the relevance of the problem in a practical setting by including experiments on data from the Gold Atlas project~\cite{nyholm2018mr} and The Lung Image Database Consortium (LIDC) and Image Database Resource
Initiative (IDRI)~\cite{armato2011lung}.

\section{Related work}
Deep learning methods are playing an increasingly vital part in the development of medical image analysis. However, deep learning models require large annotated data sets for successful training that rarely are available in the clinics.
Even if some data sets are being curated for training deep learning models, it is generally difficult and expensive to accurately annotate large collections of medical images.
Moreover, training data may include corrupted or noisy labels.
This is particularly the case in image segmentation where different annotators may have different views on the correct delineation of a region of interest, leading to uncertainty about the true label, see e.g.\ \cite{bridge2016intraobserver, nir2018automatic}. Noisy labels may also appear due to automated systems or non-expert systems being used to annotate large volumes of data, see \cite{ratner2016data, chiaroni2019hallucinating}. 

There is a large body of literature on the impact of label noise in image segmentation, see~\cite{tajbakhsh2020embracing} and \cite{karimi2020deep} for recent reviews. The proposed solutions to limit the loss of performance when the labels are noisy include label cleaning and pre-processing, e.g.\ \cite{gao2017deep}, modification of network architechtures, e.g.\ \cite{yao2018deep}, robustification of loss functions, e.g.\ \cite{matuszewski2018minimal}, reweighting of training data, see \cite{zhu2019pick,mirikharaji2019learning}, and many others. These approaches are of practical nature and generally address methodology that improve the performance on some chosen noisy data set. On the contrary, the literature that address the effect of label noise from a theoretical point of view in the context of image segmentation is rather limited.
It was shown that the loss function soft-Dice, in contrast to the loss function cross-entropy, does not yield optimal predictions that coincide with the pixel-wise marginals and that the associated volume is biased~\cite{bertels2021theoretical}. This motivated methods for post-calibrating uncalibrated marginal estimates~\cite{rousseau2021post} and a more general investigation of the relationship between volume and marginal calibration~\cite{popordanoska2021relationship}.
Finally, an alternative volume preserving segmentation method based on optimal transport theory was proposed and investigated in~\cite{liu2022deep}.

Another domain of related work can be found in the binary classification literature.
The connection between
the study of solutions to Accuracy and Dice in segmentation and the study of Accuracy and $\mathrm{F}_1$ in binary classification was investigated in~\cite{nordstrom2020calibrated}.
Of importance to us are \emph{optimal plug-in} classifiers to Accuracy and $\mathrm{F}_1$.
That is, classifiers that are obtained by processing the posterior class probabilities or estimates thereof using a threshold.
For Accuracy, this relates to the classical Bayes classifier which has been studied since the origin of the field~\cite{vapnik1999nature}.
Early work on the existence of such a classifier for $\mathrm{F}_1$, and the fact that this threshold is lower than the threshold for Accuracy can be traced to~\cite{zhao2013beyond}.
Further work showed this threshold to be equal to half of the maximal attainable $\mathrm{F}_1$-score~\cite{lipton2014optimal}.
Lots of extensions of these works have been proposed, but to the best of our knowledge no such extension is in the direction of our work.

\section{Preliminaries}
%

In our work we find that it is convenient to do the theoretical analysis over a continuous domain and where all encountered continuous spaces are equipped with their associated Borel $\sigma$-fields.
Formally, let $\Omega=[0,1]^n\subset \mathbb{R}^n$ be the unit cube of dimension $n\ge 1$ and $\lambda$ be the associated standard normalized Lebesgue measure such that $\lambda(\Omega)=1$.
The space of segmentations is denoted by $\mathcal{S}$ and is formally given by the space of measurable functions from $\Omega$ to the binary numbers $\{0,1\}$. For any segmentation $s \in \mathcal{S}$, $s(w) = 1$ implies that the object of interest occupies the site $\omega \in \Omega$.
In a numerical setting, a discretization of the domain $\Omega$ is commonly used.
For instance, 3D CT scans are often represented by a three dimensional voxel grid of an approximate order of $512\times 512 \times 100$.
In such situations the space of segmentations are given by all possible binary functions defined on this voxel grid.
Note that any segmentation on a discretized domain can be incorporated in our continuous framework by using appropriate step functions.
Details on this can be found in the end of Section 4.

Beyond the space of segmentations, several other technical constructions are introduced.
This includes the space of measurable functions from $\Omega$ to $[0,1]$ which we denote by $\mathcal{M}$ and refer to as the \emph{marginal functions}.
We also let $\lVert f\rVert_1 \doteq \int_\Omega |f(\omega)| \lambda(d\omega)$ 
and $\bar{f} \doteq 1-f$, where $f$ is any measurable function defined on $\Omega$.
Throughout we adopt the convention that two $\lambda$-measurable functions $f,g$ are equal if they are equal $\lambda$-a.e.
We will use $I\{\cdot\}$ to denote the identity function, and when $F$ is a cumulative distribution function, we will denote the left limit $F(t-) = \lim_{s \uparrow t} F(s)$.
Finally, for a given volume $v \in [0,1]$, we let $\mathcal{S}_{v} \doteq \{s\in \mathcal{S}, \lVert s\rVert_1=v\}$ be the set of segmentations with volume $v$.

Classically, metrics in medical image segmentation are defined per image as functionals over two deterministic segmentations~\cite{taha2015metrics}.
When noise is present, the label becomes a random variable and the metrics need to be extended to a functional over one deterministic segmentation and one random label segmentation.
In this work, the soft labeling convention for this extension is adopted
~\cite{kats2019soft, gros2021softseg,silva2021using,lemay2022label,li2020superpixel}.
Also, since we do our analysis with respect to measurable functions on a continuous domain instead of functions on a finite index set, we replace the sums usually used with integrals.
This, however, does not change the intuition of the metrics, and the definition using sums can be seen a special case.
\begin{definition}
For any $m\in \mathcal{M}$, Accuracy is given by
\begin{align}
    \mathrm{A}_m(s) &\doteq \int_\Omega [s(\omega)m(\omega)  + \bar{s}(\omega)\bar{m}(\omega)] \lambda(d\omega), \quad s\in\mathcal{S} 
    \label{eq:acc}
\end{align}
\end{definition}
\begin{definition}
For any $m\in \mathcal{M}$, Dice is given by
\begin{align}
    \mathrm{D}_m(s) &\doteq \frac{2\int_\Omega s(\omega)m(\omega) \lambda(d\omega)}{\lVert s\rVert_1 + \lVert m\rVert_1}, \quad s\in\mathcal{S}.\label{eq:dice}
\end{align}
\end{definition}

\begin{figure}[htb!]
  \centering
  
\subfloat{{
\includegraphics[width=0.45\textwidth]{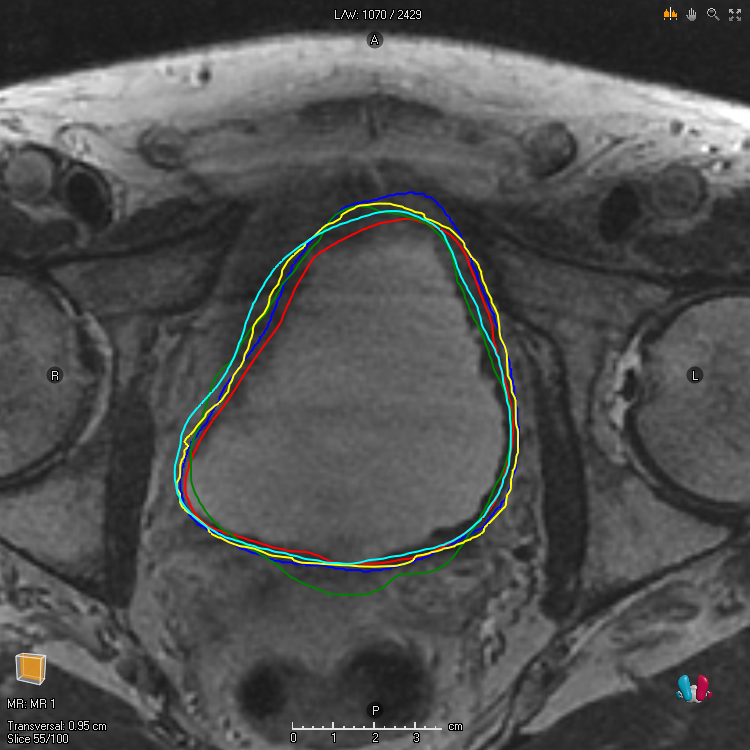}}}%
\qquad
\subfloat{{
\includegraphics[width=0.45\textwidth]{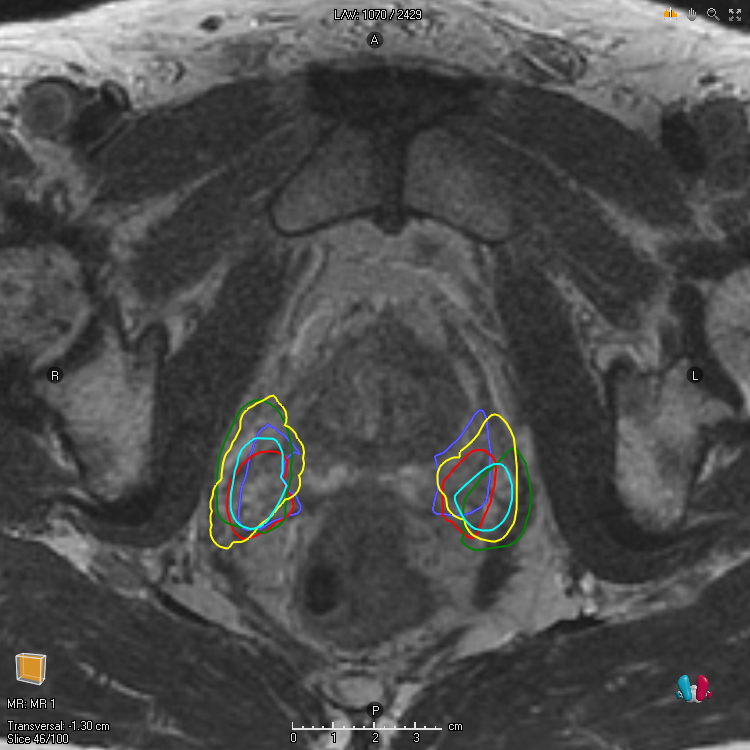}
}}%

  \caption{
  To the left is Urinary bladder and to the right is Neurovascular bundles for one patient in the Gold Atlas data~\cite{nyholm2018mr}.
  Each line is associated with the boundary of a segmentation produced by a particular annotator. The screenshots are taken with RayStation 12A (RaySearch Laboratories AB, Stockholm Sweden). }
  \label{fig:fig1}
\end{figure}

For a noisy segmentation $L$, that is, a random variable taking values in $\mathcal{S}$, $m$ can be taken to be the exact marginal success probability 
$m(\omega) = \mathbb{E}[L(\omega)],\; \omega \in \Omega$.
Such marginal functions are important in theory but can rarely be obtained in practice.
Alternative choices of marginal functions include finite sample approximations, that is, point-wise averages over finite observations of $L$, and estimates of $\mathbb{E}[L]$ according to a single annotator~\cite{kats2019soft,gros2021softseg,silva2021using,lemay2022label,li2020superpixel}.
These choices of $m$ are important because they are sometimes used for training machine learning models.
Finally, note that $\mathrm{A}_{\mathbb{E}[L]}(s) = \mathbb{E}[\mathrm{A}_L(s)]$ and $\mathbb{E}[\mathrm{A}_L(s)]$ is a common alternative way of specifying the metric.
For Dice, this sort of relationship does not hold in general $\mathrm{D}_{\mathbb{E}[L]}(s)\not= \mathbb{E}[\mathrm{D}_L(s)]$.
However, it does hold that $\mathrm{D}_{\mathbb{E}[L]}(s)= \mathbb{E}[\mathrm{D}_L(s)]$ when the volume of the noisy labels  is constant $\text{Var}[\lVert L \rVert_1]=0$, and it holds approximately
$\mathrm{D}_{\mathbb{E}[L]}(s)\approx \mathbb{E}[\mathrm{D}_L(s)]$ when the variance of the volume of the noisy labels is small $\text{Var}[\lVert L \rVert_1]\approx 0$,  which is often the case in medical image segmentation applications.
Examples of observations of a particular $L$ for a couple of different target structures are depicted in Figure~\ref{fig:fig1}.

Because of the fact that $\lVert m\rVert_1=\mathbb{E}[\lVert L\rVert_1]$ when $m$ is taken to be the exact marginal success probability, $\lVert m\rVert_1$ plays an important role in the medical segmentation context, either theoretically as the expected volume of the target or as an approximation thereof.
Understanding how $\lVert m\rVert_1$ relates to $\lVert s\rVert_1$, where $s$ is an optimal segmentation to Accuracy or Dice will be central in our work.
To the best of our knowledge, this has not been studied in prior work.

\section{Main results}
The objective of our analysis is to characterize the optimizers to $\mathrm{A}_m$ and $\mathrm{D}_m$ and give a detailed description of the volume of the optimal segmentations.
That is, for a given $m \in \mathcal{M}$, identify properties (e.g.\ volume) of the optimal segmentation $s \in \mathcal{S}$ that maximimize Accuracy or Dice.
To this end, consider the probability measure on $[0,1]$ given by the push-forward measure $\lambda \circ \bar m^{-1}(\cdot)$ and let $F_m$ denote its cumulative distribution function,
\begin{align} \label{eq:F}
F_m(t) \doteq \lambda \circ \bar m^{-1}([0,t]) = \int_\Omega I\{\bar m(\omega) \leq t\} \lambda(d\omega), \quad t\in[0,1] . 
\end{align}
The function $F_m(t)$ may be interpreted as the volume of the set of sites with non-success probability less than or equal to $t$. In other words, the volume of the sub-level set of $\bar m$ at level $t$.

\input{res_article/one_dimensional_examples_pgf}
Since $F_m$ is the cumulative distribution function of a probability distribution on $[0,1]$, it has several well-known properties making it easy to work with, e.g., $F_m$ is non-decreasing and right-continuous with $F_m(1) = 1$. Of particular interest to us is that it has a generalized inverse given by 
\begin{align}
\label{eq:Finv}
    F_m^{-1}(v) \doteq \inf\{t : F_m(t) \geq v\}, \quad v\in [0,1],
\end{align}
which can be interpreted as the minimum level at which the volume of the corresponding sub-level set of $\bar m$, is at least $v$.
This function is often referred to as the quantile function and also has several well known properties; it is non-decreasing and left-continuous.
Moreover, it allows us to define the following important class of segmentations for a given $m\in\mathcal{M}$.
\begin{align}
    \mathcal{S}_{m,v} \doteq \left\{s\in\mathcal{S}_{v} \; \middle| \;
    \begin{aligned}
    \int_\Omega  s(\omega)I\{ m(\omega) < 1-F_m^{-1}(v) \} \lambda(d\omega) = 0,  \\
    \int_\Omega  \bar{s}(\omega)I\{ m(\omega) > 1-F_m^{-1}(v)\} \lambda(d\omega) = 0,
    \end{aligned}
    \right\}, \quad v\in [0,1].
\end{align}

The described class $\mathcal{S}_{m,v}$ is informally the set of segmentations with volume $v$ that assigns $1$ to sites $\omega$ where $m(\omega)$ is large. 
If $t = F_m^{-1}(v)$ is a continuity point of $F_m$, i.e.\ $\lambda(\omega: \bar m(\omega) = t) = 0$, then $\mathcal{S}_{m,v}$ only consist of the elements that are $\lambda$-a.e.\ equal to the segmentation 
$s(\omega) = I\{ m(\omega) \ge 1-F_m^{-1}(v)\}$.
A lot of our analysis can be simplified if $F_m$ is assumed to be invertible almost everywhere.
However, this would require $m$ to not have any non-neglible constant regions, which for instance excludes any $m$ that is given by an empirical approximations using a finite number of samples.
Consequently, in the sequel, we treat general $F_m$. 
Our first result contains the essential ingredients for characterizing the optimizers to Accuracy and Dice.

\begin{lemma}
\label{lem:char} 
For any $m\in\mathcal{M}$ and $v \in [0,1]$
\begin{align} \label{eq:char}
\sup_{s\in\mathcal{S}_{v}} \int_\Omega s(\omega)  m(\omega)\lambda(d\omega)=
\int_0^v (1-F_m^{-1}(u)) du,
\end{align}
and the elements where the supremum is attained is given by $\mathcal{S}_{m,v}$.
\end{lemma}
A complete proof is given in the Supplementary Document and outlined as follows.
The first part shows that the class $\mathcal{S}_{m,v}$ is the class of optimal solutions by showing that for any $s^* \in \mathcal{S}_{m,v}$ and $s \in \mathcal{S}_v \setminus \mathcal{S}_{m,v}$, $\int_\Omega (s^*(\omega) - s(\omega))  m(\omega) \lambda(d\omega) > 0$. 
The second part proves the equality \eqref{eq:char} using an application of the quantile transform.
That is, if $U$ has uniform distribution on $[0,1]$ then $F_m^{-1}(U)$ has cdf given by $F_m$ and 
$\int_0^1 t F_m(dt) = \mathbb{E}[F_m^{-1}(U)] = \int_0^1 F_m^{-1}(u) du$.

Lemma~\ref{lem:char} allows us to reduce the constrained optimization problem over the rather complicated space of segmentations, to a one-dimensional integral with respect to the quantile function. It is the starting point for our analysis.

In the remaining section our main theoretical results are presented. In Theorem~\ref{thm:accuracy}, we provide a characterization of all of the optimal solutions to Accuracy based on volume.
In addition, sharp upper and lower bounds on the volume of the associated optimal segmentations are provided. 

\begin{figure}[htb!]
  \centering

\begin{tikzpicture}
  \begin{groupplot}[group style={group size=2 by 1, horizontal sep=2cm},height=4.5cm,width=4.5cm,xmin=-0.1,xmax=1.1,ymin=-0.1,ymax=1.1,xmajorgrids,ymajorgrids,xtick={0,0.5,1.0},ytick={0,0.5,1.0}]
    \nextgroupplot[xlabel=$v$, ylabel=$F_m^{-1}(v)$]
	\addplot[no marks,color=blue] coordinates {(0,0.5) (0.8,0.5)};
	\addplot[no marks,color=blue] coordinates {(0.8,1.0) (1.0,1.0)};
    \addplot[only marks,mark=*,color=blue,fill=blue] coordinates{(0,0)};
    \addplot[only marks,mark=*,color=blue,fill=white] coordinates{(0,0.5)};
    \addplot[only marks,mark=*,color=blue,fill=blue] coordinates{(0.8,0.5)};
    \addplot[only marks,mark=*,color=blue,fill=white] coordinates{(0.8,1.0)};
    \addplot[only marks,mark=*,color=blue,fill=blue] coordinates{(1.0,1.0)};
    
    \nextgroupplot[xlabel=$v$, ylabel=$a_m(v)$]
    \addplot[no marks,color=blue]
    coordinates{
    (0.0,0.6)(0.01001001001001001,0.6)(0.02002002002002002,0.6)(0.03003003003003003,0.6)(0.04004004004004004,0.6)(0.050050050050050046,0.6)(0.06006006006006006,0.6)(0.07007007007007007,0.6)(0.08008008008008008,0.6)(0.09009009009009009,0.6)(0.10010010010010009,0.6)(0.11011011011011011,0.6)(0.12012012012012012,0.6)(0.13013013013013014,0.5999999999999999)(0.14014014014014015,0.5999999999999999)(0.15015015015015015,0.5999999999999999)(0.16016016016016016,0.5999999999999999)(0.17017017017017017,0.5999999999999999)(0.18018018018018017,0.5999999999999999)(0.19019019019019018,0.5999999999999999)(0.20020020020020018,0.5999999999999999)(0.21021021021021022,0.5999999999999999)(0.22022022022022023,0.5999999999999999)(0.23023023023023023,0.5999999999999999)(0.24024024024024024,0.5999999999999999)(0.2502502502502503,0.5999999999999999)(0.2602602602602603,0.5999999999999999)(0.2702702702702703,0.5999999999999999)(0.2802802802802803,0.5999999999999999)(0.2902902902902903,0.5999999999999999)(0.3003003003003003,0.5999999999999999)(0.3103103103103103,0.5999999999999999)(0.3203203203203203,0.5999999999999999)(0.3303303303303303,0.5999999999999999)(0.34034034034034033,0.5999999999999999)(0.35035035035035034,0.5999999999999999)(0.36036036036036034,0.5999999999999999)(0.37037037037037035,0.5999999999999999)(0.38038038038038036,0.5999999999999999)(0.39039039039039036,0.5999999999999999)(0.40040040040040037,0.6)(0.41041041041041043,0.5999999999999999)(0.42042042042042044,0.5999999999999999)(0.43043043043043044,0.5999999999999999)(0.44044044044044045,0.5999999999999999)(0.45045045045045046,0.5999999999999999)(0.46046046046046046,0.5999999999999999)(0.47047047047047047,0.5999999999999999)(0.4804804804804805,0.5999999999999999)(0.4904904904904905,0.5999999999999999)(0.5005005005005005,0.5999999999999999)(0.5105105105105106,0.5999999999999999)(0.5205205205205206,0.5999999999999999)(0.5305305305305306,0.5999999999999999)(0.5405405405405406,0.5999999999999999)(0.5505505505505506,0.5999999999999999)(0.5605605605605606,0.5999999999999999)(0.5705705705705706,0.5999999999999999)(0.5805805805805806,0.5999999999999999)(0.5905905905905906,0.5999999999999999)(0.6006006006006006,0.5999999999999999)(0.6106106106106106,0.5999999999999999)(0.6206206206206206,0.5999999999999999)(0.6306306306306306,0.5999999999999999)(0.6406406406406406,0.5999999999999999)(0.6506506506506506,0.5999999999999999)(0.6606606606606606,0.5999999999999999)(0.6706706706706707,0.5999999999999999)(0.6806806806806807,0.5999999999999999)(0.6906906906906907,0.5999999999999999)(0.7007007007007007,0.5999999999999999)(0.7107107107107107,0.5999999999999999)(0.7207207207207207,0.5999999999999999)(0.7307307307307307,0.5999999999999999)(0.7407407407407407,0.5999999999999999)(0.7507507507507507,0.5999999999999999)(0.7607607607607607,0.5999999999999999)(0.7707707707707707,0.5999999999999999)(0.7807807807807807,0.5999999999999999)(0.7907907907907907,0.5999999999999999)(0.8008008008008007,0.5989989989989988)(0.8108108108108109,0.5889889889889892)(0.8208208208208209,0.5789789789789792)(0.8308308308308309,0.5689689689689692)(0.8408408408408409,0.5589589589589592)(0.8508508508508509,0.5489489489489492)(0.8608608608608609,0.5389389389389392)(0.8708708708708709,0.5289289289289292)(0.8808808808808809,0.5189189189189192)(0.8908908908908909,0.5089089089089092)(0.9009009009009009,0.49889889889889916)(0.9109109109109109,0.48888888888888915)(0.9209209209209209,0.47887887887887914)(0.9309309309309309,0.46886886886886914)(0.9409409409409409,0.45885885885885913)(0.950950950950951,0.4488488488488491)(0.960960960960961,0.4388388388388391)(0.970970970970971,0.4288288288288291)(0.980980980980981,0.4188188188188191)(0.990990990990991,0.4088088088088091)(1.0,0.39979979979979996)
    };
    \addplot[only marks,mark=*,color=blue,fill=blue] coordinates{(0.0,0.6)};
    \addplot[only marks,mark=*,color=blue,fill=blue] coordinates{(1.0,0.39979979979979996)};
  \end{groupplot}
\end{tikzpicture}

  \caption{To the left is a particular quantile function $F_m^{-1}$ and to the right is the associated function $a_m$ given by \eqref{eq:a}. Here $F_{m}^{-1}(v) = \frac{1}{2} I_{(0,2\lVert m\rVert_1]}(v) + I_{(2\lVert m\rVert_1, 1]}(v), v \in [0,1]$ with $\lVert m\rVert_1 = 0.4$ that satisfies $\mathcal{V}^{\mathrm{A}_m} = [0,2\lVert m\rVert_1] = [\max\{2 \lVert m\rVert_1-1,0\},\min\{2\lVert m\rVert_1,1\}]$ and consequently is an extreme case to \eqref{sharpacc} in both the lower sense and the upper sense.
  }
  \label{fig:extremeacc}
\end{figure}
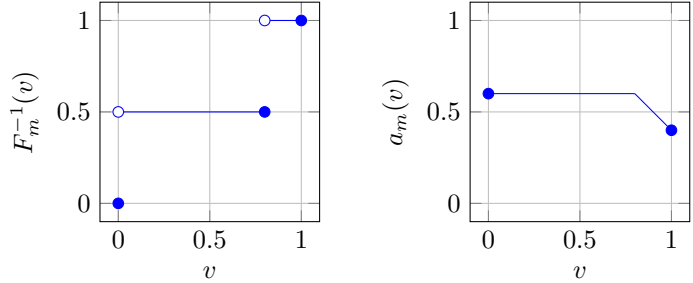

\begin{theorem}
\label{thm:accuracy}
For any $m\in\mathcal{M}$, the class of maximizers to $\mathrm{A}_m$ is given by $\cup_{v\in\mathcal{V}^{\mathrm{A}_m}} \mathcal{S}_{m,v}$ where
\begin{align}
    \mathcal{V}^{\mathrm{A}_m} \doteq [F_m(1/2-), F_m(1/2)] \label{characc}.
\end{align}
Moreover, $\mathcal{V}^{\mathrm{A}_m}$ satisfies the following bounds
\begin{align}
    \mathcal{V}^{\mathrm{A}_m} \subseteq [\max\{ 2\lVert m\rVert_1-1, 0 \},\min\{ 2\lVert m\rVert_1,1\}] \label{boundsacc},
\end{align}
where the bounds are sharp in the sense that there for any $v\in[0,1]$ exist $m_0,m_1\in\mathcal{M}$ such that $\lVert m_0\rVert_1=\lVert m_1\rVert_1=v$ and
\begin{align}
    \inf \mathcal{V}^{\mathrm{A}_{m_0}} = \max\{ 2\lVert m_0\rVert_1-1, 0 \},\quad
    \sup \mathcal{V}^{\mathrm{A}_{m_1}} =  \min\{ 2\lVert m_1\rVert_1,1\}. \label{sharpacc}
\end{align}
\end{theorem}
The complete proof is given in the Supplementary Document and  outlined as follows. First, the function
\begin{align}
    a_m(v) \doteq  v+1-\lVert m\rVert_1 - 2\int_0^v F_m^{-1}(u) du, \quad v\in[0,1], \label{eq:a}
\end{align} 
is introduced and then Lemma~\ref{lem:char} is used to show that
$\sup_{s\in\mathcal{S}} \mathrm{A}_m(s) =\sup_{v\in[0,1]} a_m(v)$.  Consequently, the class of optimal solutions to $\mathrm{A}_m$ is given by $\cup_{v\in\mathcal{V}^{\mathrm{A}_m}}\mathcal{S}_{m,v}$, where $\mathcal{V}^{\mathrm{A}_m}$ is the set of optimizers to $a_m$.
The rest of the proof consists of detailed analysis of $a_m$ and  is composed of three parts. The first part is to show~\eqref{characc} by finding one optimal solution and then identifying all volumes that yield the same optimal value. The second part is to provide the lower and upper bounds on the elements of $\mathcal{V}^{\mathrm{A}_m}$ in terms of $\lVert m\rVert_1$ given by~\eqref{boundsacc}. The third part is to provide examples of situations where the extreme cases occur~\eqref{sharpacc}. In Figure~\ref{fig:extremeacc},
a case that is extreme both in the lower and in the upper sense is illustrated.

In Theorem~\ref{thm:dice}, we provide a characterization of all of the optimal solutions to Dice based on volume.
In addition, sharp upper and lower bounds on the volume of the associated optimal segmentations are provided.

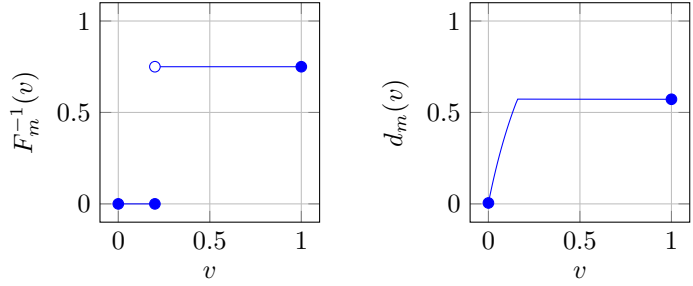
\begin{figure}[htb!]
  \centering

\begin{tikzpicture}
  \begin{groupplot}[group style={group size=2 by 1, horizontal sep=2cm},height=4.5cm,width=4.5cm,xmin=-0.1,xmax=1.1,ymin=-0.1,ymax=1.1,xmajorgrids,ymajorgrids,xtick={0,0.5,1.0},ytick={0,0.5,1.0}]
    \nextgroupplot[xlabel=$v$, ylabel=$F_m^{-1}(v)$]
	\addplot[no marks,color=blue] coordinates {(0,0) (0.2,0.0)};
	\addplot[no marks,color=blue] coordinates {(0.2,0.75) (1.0,0.75)};
	
    \addplot[only marks,mark=*,color=blue,fill=blue] coordinates{(0,0)};
    \addplot[only marks,mark=*,color=blue,fill=blue] coordinates{(0.2,0.0)};
    \addplot[only marks,mark=*,color=blue,fill=white] coordinates{(0.2,0.75)};
    \addplot[only marks,mark=*,color=blue,fill=blue] coordinates{(1.0,0.75)};
    
    \nextgroupplot[xlabel=$v$, ylabel=$d_m(v)$]
    \addplot[no marks,color=blue]
    coordinates{
    (0.0,0.005005005005005005)(0.01001001001001001,0.0537109375)(0.02002002002002002,0.10009532888465204)(0.03003003003003003,0.14432029795158288)(0.04004004004004004,0.186533212010919)(0.050050050050050046,0.22686832740213522)(0.06006006006006006,0.2654482158398607)(0.07007007007007007,0.3023850085178875)(0.08008008008008008,0.3377814845704754)(0.09009009009009009,0.37173202614379086)(0.10010010010010009,0.4043234587670136)(0.11011011011011011,0.4356357927786499)(0.12012012012012012,0.4657428791377983)(0.13013013013013014,0.49471299093655574)(0.14014014014014015,0.5226093402520385)(0.15015015015015015,0.5494905385735079)(0.16016016016016016,0.57285816399469)(0.17017017017017017,0.5728330658105935)(0.18018018018018017,0.5728088336783983)(0.19019019019019018,0.5727854235316913)(0.20020020020020018,0.5727627942437804)(0.21021021021021022,0.5727409073865752)(0.22022022022022023,0.5727197270128177)(0.23023023023023023,0.5726992194590651)(0.24024024024024024,0.5726793531671559)(0.2502502502502503,0.5726600985221652)(0.2602602602602603,0.5726414277051002)(0.2702702702702703,0.5726233145587958)(0.2802802802802803,0.5726057344656491)(0.2902902902902903,0.572588664235993)(0.3003003003003003,0.5725720820060409)(0.3103103103103103,0.5725559671444642)(0.3203203203203203,0.5725403001667556)(0.3303303303303303,0.5725250626566378)(0.34034034034034033,0.572510237193846)(0.35035035035035034,0.5724958072876921)(0.36036036036036034,0.5724817573158762)(0.37037037037037035,0.572468072468068)(0.38038038038038036,0.572454738693831)(0.39039039039039036,0.5724417426545039)(0.40040040040040037,0.5724290716786916)(0.41041041041041043,0.5724167137210565)(0.42042042042042044,0.5724046573241252)(0.43043043043043044,0.5723928915828573)(0.44044044044044045,0.5723814061117485)(0.45045045045045046,0.5723701910142532)(0.46046046046046046,0.5723592368543452)(0.47047047047047047,0.5723485346300375)(0.4804804804804805,0.572338075748711)(0.4904904904904905,0.572327852004105)(0.5005005005005005,0.5723178555548436)(0.5105105105105106,0.5723080789043786)(0.5205205205205206,0.5722985148822405)(0.5305305305305306,0.5722891566264995)(0.5405405405405406,0.5722799975673473)(0.5505505505505506,0.5722710314117154)(0.5605605605605606,0.5722622521288564)(0.5705705705705706,0.5722536539368154)(0.5805805805805806,0.5722452312897323)(0.5905905905905906,0.5722369788659127)(0.6006006006006006,0.5722288915566155)(0.6106106106106106,0.5722209644555055)(0.6206206206206206,0.5722131928487287)(0.6306306306306306,0.5722055722055647)(0.6406406406406406,0.5721980981696181)(0.6506506506506506,0.5721907665505149)(0.6606606606606606,0.5721835733160684)(0.6706706706706707,0.5721765145848837)(0.6806806806806807,0.572169586619375)(0.6906906906906907,0.5721627858191655)(0.7007007007007007,0.5721561087148492)(0.7107107107107107,0.5721495519620889)(0.7207207207207207,0.5721431123360302)(0.7307307307307307,0.5721367867260135)(0.7407407407407407,0.5721305721305637)(0.7507507507507507,0.572124465652641)(0.7607607607607607,0.5721184644951377)(0.7707707707707707,0.5721125659566066)(0.7807807807807807,0.5721067674272065)(0.7907907907907907,0.5721010663848506)(0.8008008008008007,0.5720954603915503)(0.8108108108108109,0.5720899470899383)(0.8208208208208209,0.5720845241999631)(0.8308308308308309,0.5720791895157453)(0.8408408408408409,0.5720739409025863)(0.8508508508508509,0.5720687762941193)(0.8608608608608609,0.5720636936895976)(0.8708708708708709,0.5720586911513097)(0.8808808808808809,0.5720537668021165)(0.8908908908908909,0.5720489188231032)(0.9009009009009009,0.5720441454513384)(0.9109109109109109,0.5720394449777374)(0.9209209209209209,0.5720348157450216)(0.9309309309309309,0.5720302561457702)(0.9409409409409409,0.5720257646205595)(0.950950950950951,0.5720213396561848)(0.960960960960961,0.5720169797839605)(0.970970970970971,0.5720126835780945)(0.980980980980981,0.5720084496541344)(0.990990990990991,0.5720042766674793)(1.0,0.5720005720005622)
    };
    \addplot[only marks,mark=*,color=blue,fill=blue] coordinates{(0.0,0.005005005005005005)};
    \addplot[only marks,mark=*,color=blue,fill=blue] coordinates{(1.0,0.5720005720005622)};
  \end{groupplot}
\end{tikzpicture}
  \caption{To the left is a particular quantile function $F_m^{-1}$ and to the right is the associated function $d_m$ given by \eqref{eq:d}. Here $F^{-1}_m(v)=(1-\lVert m\rVert_1)(1-\lVert m\rVert_1^2)^{-1}I_{ (\lVert m\rVert_1^2,1] }(v)$ with $\lVert m\rVert_1 = 0.4$ that satisfies $\mathcal{V}^{\mathrm{D}_m} = [\lVert m\rVert_1^2,1]$ and consequently is an extreme case to \eqref{sharpdice} in both the lower sense and the upper sense.
  }
  \label{fig:extremedice}
\end{figure}

\begin{theorem}
\label{thm:dice}
For any $m\in\mathcal{M}$, the class of maximizers to $\mathrm{D}_m$ is given by 
$\cup_{v\in\mathcal{V}^{\mathrm{D}_m}} \mathcal{S}_{m,v}$ where
\begin{align}
    \mathcal{V}^{\mathrm{D}_m}\doteq [F_m((1-\sup_{s\in \mathcal{S}} \mathrm{D}_m(s)/2)-), F_m(1-\sup_{s\in \mathcal{S}}\mathrm{D}_m(s)/2)] \label{chardice}.
\end{align}
Moreover, $\mathcal{V}^{\mathrm{D}_m}$ satisfies the following bounds 
\begin{align}
    \mathcal{V}^{\mathrm{D}_m} \subseteq [\lVert m\rVert_1^2, 1]\label{boundsdice},
\end{align}
where the bounds are sharp in the sense that there for any $v\in (0,1]$ exist $m_0,m_1\in\mathcal{M}$ such that $\lVert m_0\rVert_1=\lVert m_1\rVert_1=v$ and
\begin{align}
    \inf \mathcal{V}^{\mathrm{D}_{m_0}} = \lVert m\rVert_1^2,\quad
    \sup \mathcal{V}^{\mathrm{D}_{m_1}} =  1. \label{sharpdice}
\end{align}
\end{theorem}
The complete proof is given in the Supplementary Document and outlined as follows. First, the function
\begin{align}
    d_m(v) \doteq \frac{2\int_0^v (1-F_m^{-1}(u) )du}{\lVert m\rVert_1+v}, \quad v\in[0,1], \label{eq:d}
\end{align}
is introduced  and then Lemma~\ref{lem:char} is used to show that
$\sup_{s\in\mathcal{S}} \mathrm{D}_m(s) =\sup_{v\in[0,1]} d_m(v)$. 
Consequently, the class of optimal solutions to $\mathrm{D}_m$ is given by
$\cup_{v\in\mathcal{V}^{\mathrm{D}_m}}\mathcal{S}_{m,v}$, where $\mathcal{V}^{D_m}$ is the set of optimizers to $d_m$.
The remaining proof consists of detailed analysis of $d_m$ and composed of three parts. The first part is to show~\eqref{chardice} which is derived by careful investigation of the properties of the function  $\delta(v) = \frac{(\lVert m\rVert_1+v)^2}{2}\partial_v d_m(v)$, which has the same sign as $\partial_v d_m(v)$ and therefore can be used to identify optimal values of $d_m$. The second part is to provide the lower and upper bounds on the elements of $\mathcal{V}^{\mathrm{D}_m}$ in terms of $\lVert m\rVert_1$ given by \eqref{boundsdice}.
The third part is to provide examples of situations where the extreme values occur~\eqref{sharpdice}. 
In Figure~\ref{fig:extremedice}, a case that is extreme both in the lower and in the upper sense is illustrated.

In Theorem~\ref{thm:relation}, we relate the volume of the optimal segmentations of Accuracy and the optimal segmentations of Dice for a given marginal probability $m \in \mathcal{M}$.
\begin{theorem}
\label{thm:relation}
For any $m\in\mathcal{M}$, $\mathcal{V}^{\mathrm{A}_m}$ given by~\eqref{characc} and $\mathcal{V}^{\mathrm{D}_m}$ given by~\eqref{chardice} satisfy
\begin{align}
    \sup \mathcal{V}^{\mathrm{A}_m} \le \inf \mathcal{V}^{\mathrm{D}_m} \label{releq}.
\end{align}
\end{theorem}
The complete proof is given in the Supplementary Document and outlined as follows. First note that $\mathrm{D}_m(s) \le 1$ for any $s\in\mathcal{S}$ and then consider separately the cases when $\mathrm{D}_m(s)<1$ for all $s\in\mathcal{S}$ and when there exist some $s\in\mathcal{S}$ such that $\mathrm{D}_m(s)=1$.
For the first case, it is obvious that $\sup_{s\in\mathcal{S}} \mathrm{D}_m(s)/2 < 1/2$ which implies that $F_m(1/2) \le F_m((1-\sup_{s\in\mathcal{S}} \mathrm{D}_m(s)/2)-) $. For the second case, we show that the volume of the optimizers are uniquely given by $\mathcal{V}^{\mathrm{A}_m}=\mathcal{V}^{\mathrm{D}_m}=\{F_m(1/2)\}$.
In either case, \eqref{releq} holds.

In Theorem~\ref{thm:constrained}, the set of optimal solutions to Accuracy and Dice when constrained to a specific volume is  shown to coincide.
\begin{theorem}
\label{thm:constrained}
For any $m\in\mathcal{M}$ and $v \in [0,1]$ the maximizers to the problems,
\begin{align}
\sup_{s\in\mathcal{S}_{v}} \mathrm{A}_m(s) \quad \text{  and  }\quad
\sup_{s\in\mathcal{S}_{v}} \mathrm{D}_m(s), \label{eq:volcon}
\end{align}
coincide and are given by $\mathcal{S}_{m,v}$.
\end{theorem}
The complete proof is given in the Supplementary Document and is a straightforward application of~Lemma~\ref{lem:char}.
Of particular interest is the case when $v=\lVert m\rVert_1$, since this correspond to the situation when the  metrics are maximized under the constraint that there should be no volume bias.

It follows from Theorem \ref{thm:accuracy} and Theorem \ref{thm:dice} that the optimizers to both Accuracy and Dice are of the form $\cup_{v\in\mathcal{V}} \mathcal{S}_{m,v}$, where $\mathcal{V} = [F_m(t-),F_m(t)]$ for some $t\in [0,1]$.
This type of charecterization is practical for proving properties on volume, but inconvenient for other tasks.
In Theorem~\ref{thm:threshold}, we provide an alternative charecterization using
threshold segmentations of the form $s(\omega) = I\{m(\omega) > \alpha\}$ or $s(\omega)=I\{ m(\omega) \ge \alpha\}$, for some $\alpha$.
Even if there exist optimal segmentations that are not necessarily of threshold type, they can always be bounded, above and below, by optimal segmentations of threshold type. 
\begin{theorem}
\label{thm:threshold}
For any $m\in\mathcal{M}$ and $t \in (0,1]$, let $s_1(\omega) = I\{m(\omega) \ge 1-t\}$ and $s_0(\omega) = I\{m(\omega) > 1-t\}$. Then, $\lVert s_0\rVert_1 = F_m(t-)$, $\lVert s_1\rVert_1 = F_m(t)$ and 
\begin{align}
s \in \cup_{v\in [F_m(t-),F_m(t)]}\mathcal{S}_{m,v} \quad \Longleftrightarrow \quad
s_0(\omega) \le s(\omega) \le s_1(\omega), \quad  \lambda-{a.e.} \label{threshiff}
\end{align}
\end{theorem}
The complete proof is given in the Supplementary Document and outlined as follows.
Note that $\lVert s_0\rVert_1 = \int_\Omega I\{\bar m(\omega) < t\}\lambda(d\omega) = F_m(t-)$ and 
$\lVert s_1\rVert_1 = \int_\Omega I\{\bar m(\omega) \leq  t\}\lambda(d\omega) = F_m(t)$.
Now, take each direction of \eqref{threshiff} separately.
For the $\Rightarrow$ part, we first show the upper bound $s(\omega)\le s_1(\omega),\lambda$-a.e. and then show the lower bound $s_0(\omega) \le s(\omega), \lambda$-a.e.
For the upper bound, with  $A = \{\omega : s(\omega) > s_1(\omega)\}$, we first observe that $I\{\omega \in A\} = s(\omega)\bar s_1(\omega)$ and then, using the definition of $\mathcal{S}_{m,v}$ we prove that 
\begin{align}
    \lambda(A) = \int_\Omega s(\omega) \bar s_1(\omega) \lambda(d\omega) 
    \leq \int_\Omega s(\omega) I\{\bar m(\omega) > F_m^{-1}(v)\} \lambda(d\omega)
    = 0.
\end{align}
The lower bound is similar, but slightly more involved. 
For the $\Rightarrow$ part, we first note that the $F_m(t-)=\lVert s_0\rVert_1\le \lVert s \rVert_1 \le \lVert s_1 \rVert_1 = F_m(t)$, and then show that for $v=\lVert s \rVert_1$, $s\in\mathcal{S}_{m,v}$.

In numerical applications, the continuum $\Omega$ is usually partitioned into a finite collection of voxels $\{ \Omega_i \}_{i\in\mathcal{I}}$.
Marginal functions are then constrained to the subset of $\mathcal{M}$ that is compatible with the voxelization in the sense that $m$ is measurable with respect to the $\sigma$-field generated by the partition. 
Note that if $s_1(\omega) = I\{m(\omega) \ge 1-t\}$ and $s_0(\omega) =I\{ m(\omega) > 1-t \}$ for some $t\in (0,1]$, then also
$s_0$ and $s_1$ are compatible with the voxelization. 
By Theorem~\ref{thm:accuracy} (Theorem~\ref{thm:dice}) and Theorem~\ref{thm:threshold}, the segmentations with least and greatest volume
that are optimal with respect to Accuracy (Dice) are compatible with the voxelization.
%
For the metrics respectively, we denote the segmentations with the greatest volume by:
\begin{align}
    s^{\mathrm{A}_m}(\omega) &\doteq I\{ m(\omega) \ge 1/2 \}, \quad \omega \in \Omega, \label{eq:sa} \\
    s^{\mathrm{D}_m}(\omega) &\doteq I\{m(\omega) \ge \sup_{s\in\mathcal{S}} \mathrm{D}_m(s)/2 \}, \quad \omega \in \Omega. \label{eq:sd}
\end{align}
Note that  $s^{\mathrm{A}_m}$ is analogous to the Bayes classifier and $s^{\mathrm{D}_m}$ is analogous to the threshold classifier described in~\cite{lipton2014optimal}.
Both of these are  trivial to compute from a given marginal function $m$ compatible to some voxelization and code for doing so is available in the Supplementary Material.

\section{Experiments}
The sharp bounds on volume in Thereom~\ref{thm:accuracy} and Theorem~\ref{thm:dice} implies that there exist marginal functions for which the volume of the optimal segmentations to Accuracy and Dice deviate significantly from  the expected target volume. In this section
we conduct experiments on marginal functions formed from real world data to compare the volume of optimal segmentations to the expected volume in practice.

For our experiments we investigate two data sets. The first data set (G) contains segmentations in the pelvic area and is part of the Gold Atlas project~\cite{nyholm2018mr}.
The data is in 3D with a resolution of $512\times 512$ pixels per slice and consist of $19$ patients with $9$ different ROI's (region of interest), each of which have been delineated by $5$ experts (see Figure~\ref{fig:fig1} for an illustration of the segmentations associated with two different ROI's for one patient).
The second data set (L) contains segmentations in the thorax area and is part of The Lung Image Database Consortium (LIDC) and Image Database Resource Initiative (IDRI)~\cite{armato2011lung} and is hosted by TCIA~\cite{clark2013cancer}.
The data is in 3D with a resolution of $512\times 512$ pixels per slice and contains $1018$ cases with lung nodules delineated by $4$ experts.
For each data set, ROI and patient, a marginal function $m$ is formed by taking the fraction of which each pixel has been selected by the annotators, that is, a finite sample approximation is considered.
The resulting marginal functions are then used to compute the segmentations
$s^{\mathrm{A}_m}$ \eqref{eq:sa} and $s^{\mathrm{D}_m}$ \eqref{eq:sd}.
For (G) we make use of the software \emph{Plastimatch}~\cite{sharp2010plastimatch} and for (L) we make use of the python package \emph{pylidc}~\cite{hancock2016lung}. Details on the experiments can be found in the Supplementary Document.
Code and instructions on how to reproduce the experiments can be found at
\url{https://github.com/marcus-nordstrom/optimal-solutions-to-accuracy-and-dice}.

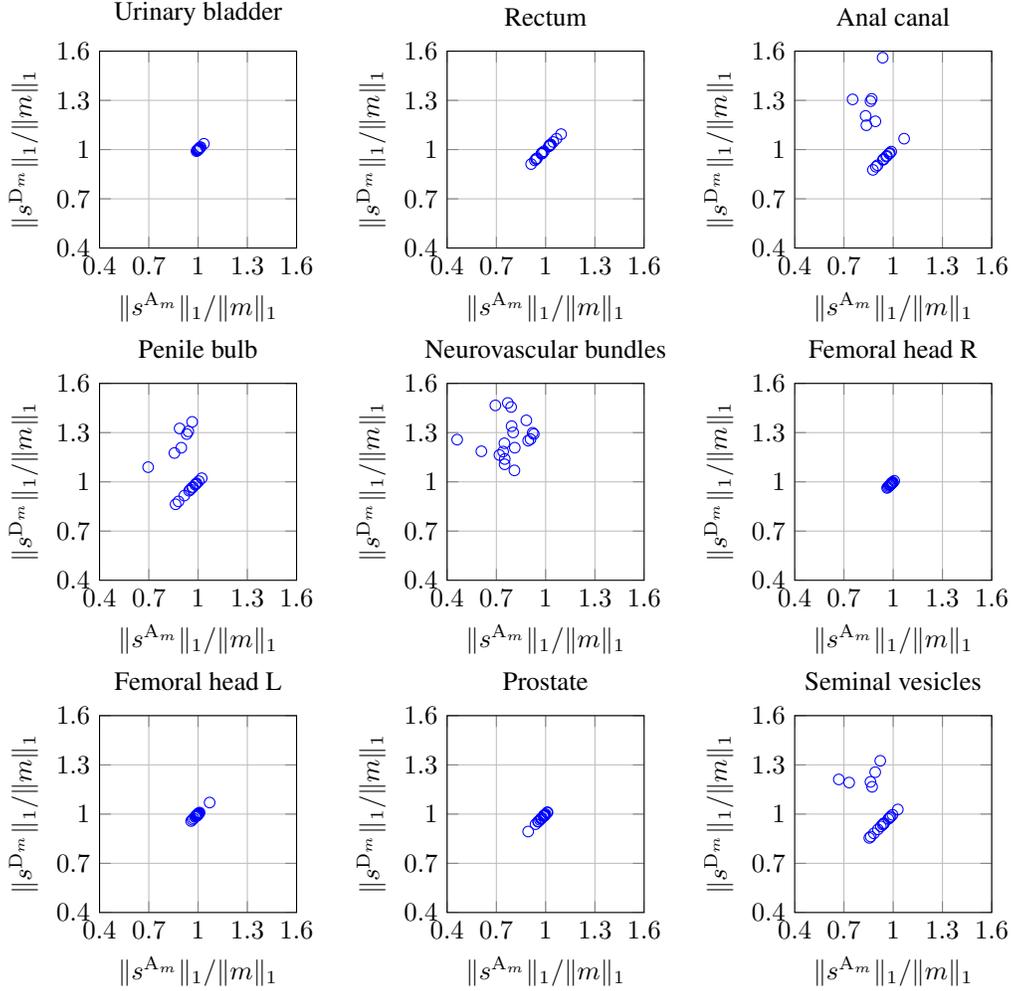
\begin{figure}[htb!]
  \centering

\begin{tikzpicture}
  \begin{groupplot}[group style={group size=3 by 3, horizontal sep=2cm, vertical sep=1.8cm},height=4.2cm,width=4.2cm,xmajorgrids,ymajorgrids,xtick={0.4,0.7,1.0,1.3,1.6},xmin=0.4,xmax=1.6,ytick={0.4,0.7,1.0,1.3,1.6},ymin=0.4,ymax=1.6,xlabel=$\lVert s^{\mathrm{A}_m}\rVert_1/\lVert m\rVert_1$, ylabel=$\lVert s^{\mathrm{D}_m}\rVert_1/\lVert m\rVert_1$]
  
    \nextgroupplot[title=Urinary bladder]
	\addplot[only marks,mark=*,color=blue,fill opacity=0.0] coordinates
	{
	(1.0022253276728768,1.0022253276728768)(1.0018925130405547,1.0018925130405547)(0.9951143015659147,0.9951143015659147)(1.0353677907415237,1.0353677907415237)(1.0071888971273772,1.0071888971273772)(1.004118809836968,1.004118809836968)(1.013418652943301,1.013418652943301)(0.9994336144443392,0.9994336144443392)(1.007438128922935,1.007438128922935)(1.0009957612459353,1.0009957612459353)(1.00219420811857,1.00219420811857)(1.0148752876574239,1.0148752876574239)(0.9974337698489703,0.9974337698489703)(0.9986504390512438,0.9986504390512438)(0.9960138923802241,0.9960138923802241)(0.9906175512568477,0.9906175512568477)(1.004243011743767,1.004243011743767)(1.0054508421630293,1.0054508421630293)(1.0014785271776412,1.0014785271776412)
	};
    
    \nextgroupplot[title=Rectum]
	\addplot[only marks,mark=*,color=blue,fill opacity=0.0] coordinates
	{
	(0.9769668622865464,0.9769668622865464)(0.9807735027042216,0.9807735027042216)(0.9476495429666878,0.9476495429666878)(1.0299585036019714,1.0299585036019714)(1.0308226874340936,1.0308226874340936)(0.9898449542116239,0.9898449542116239)(0.9343537928999618,0.9343537928999618)(1.0195403152848015,1.0195403152848015)(1.02257072530572,1.02257072530572)(1.0661084232343423,1.0661084232343423)(0.9785415289711392,0.9785415289711392)(0.9788637154575052,0.9788637154575052)(1.046298362576361,1.046298362576361)(1.0941462587966664,1.0941462587966664)(0.9430689165746728,0.9430689165746728)(0.9745001421692845,0.9745001421692845)(1.0263960810940762,1.0263960810940762)(0.9118747873426334,0.9118747873426334)(0.9427319314822705,0.9427319314822705)
	};
	
    \nextgroupplot[title=Anal canal]
	\addplot[only marks,mark=*,color=blue,fill opacity=0.0] coordinates
	{
	(0.8369807871469,1.1483766633268293)(0.9618282270215971,0.9618282270215971)(0.7528319880905309,1.3059710171896448)(0.9045718519226629,0.9045718519226629)(0.9792439703153991,0.9792439703153991)(0.9878806979342749,0.9878806979342749)(0.8692370010310796,1.3091029606716746)(1.0665688910616802,1.0665688910616802)(0.9426471207717262,0.9426471207717262)(0.9373534748129572,0.9373534748129572)(0.8765061317887659,0.8765061317887659)(0.8951662725549091,0.8951662725549091)(0.8917053495366748,1.1723082807420155)(0.9752027593921878,0.9752027593921878)(0.8620417014932117,1.2948697675469312)(0.9405858606769487,0.9405858606769487)(0.9575880159281966,0.9575880159281966)(0.9358123996561953,1.5600116763699456)(0.8320561002750424,1.2053580801612462)
	};

    \nextgroupplot[title=Penile bulb]
	\addplot[only marks,mark=*,color=blue,fill opacity=0.0] coordinates
	{
	(0.6960008789277081,1.08959569325423)(1.004257818235208,1.004257818235208)(0.9638839861963594,0.9638839861963594)(0.8560362371167858,1.1757368870215976)(0.9161740999462654,0.9161740999462654)(0.8808486034513433,0.8808486034513433)(0.9640522875816994,1.365286855482934)(0.9406006085461038,1.3077126604048153)(1.0221810534748648,1.0221810534748648)(0.991052830903238,0.991052830903238)(0.9682935847661304,0.9682935847661304)(0.9308525724658235,1.2913381360258667)(0.8978145304193736,1.2079149438865915)(0.9823411664701529,0.9823411664701529)(0.9456710119810111,0.9456710119810111)(0.8632347291586631,0.8632347291586631)(0.9529790013061388,0.9529790013061388)(0.887577718563634,1.3253394239309733)(0.9871631258914495,0.9871631258914495)
	};
    
    \nextgroupplot[title=Neurovascular bundles]
	\addplot[only marks,mark=*,color=blue,fill opacity=0.0] coordinates
	{
	(0.6087144089732527,1.18622375611159)(0.7411107906555147,1.184438218450002)(0.7185884501524366,1.1636007481921142)(0.6939847460066194,1.4658583968916385)(0.8020884133222408,1.3009552769729478)(0.9210749262936728,1.298925751465654)(0.7512487005907554,1.1390431277097433)(0.7897555296856809,1.456228172293364)(0.9080242240726724,1.2600302800908405)(0.8099409344384054,1.0697822473208283)(0.7502688002023906,1.1069824805515145)(0.4612546125461256,1.2572846418119359)(0.812058272032309,1.2080709008445913)(0.7927923352969058,1.3393763965061958)(0.7489961906722945,1.2349428600844223)(0.9279888225964801,1.2910184276112069)(0.8825206729398347,1.3747160057764656)(0.7696524281305833,1.4806318012018844)(0.8937809266756636,1.2511246063877643)
	};
    
    \nextgroupplot[title=Femoral head R]
	\addplot[only marks,mark=*,color=blue,fill opacity=0.0] coordinates
	{
	(0.9948631393556909,0.9948631393556909)(0.9962665150385422,0.9962665150385422)(0.989116995411131,0.989116995411131)(0.974327527991872,0.974327527991872)(0.9977201431012007,0.9977201431012007)(0.9946329903564401,0.9946329903564401)(0.9820880420200551,0.9820880420200551)(0.9796851249847446,0.9796851249847446)(0.9628574378032422,0.9628574378032422)(0.9814269877689916,0.9814269877689916)(0.9959275612316492,0.9959275612316492)(0.9908639780547102,0.9908639780547102)(0.9926415304840486,0.9926415304840486)(0.9835003473835415,0.9835003473835415)(0.9680450359912091,0.9680450359912091)(0.9936333559359054,0.9936333559359054)(0.989252832596418,0.989252832596418)(0.9897743375750372,0.9897743375750372)(1.0062226768470146,1.0062226768470146)
	};
    
    \nextgroupplot[title=Femoral head L]
	\addplot[only marks,mark=*,color=blue,fill opacity=0.0] coordinates
	{
	(0.9948179340651477,0.9948179340651477)(1.0022606569966797,1.0022606569966797)(0.99891639336157,0.99891639336157)(0.9943734294766741,0.9943734294766741)(0.9875579855991219,0.9875579855991219)(1.0049223494680481,1.0049223494680481)(0.9912723672998847,0.9912723672998847)(0.986302927248821,0.986302927248821)(0.9651284618092819,0.9651284618092819)(0.9904696087618813,0.9904696087618813)(1.0037070004845015,1.0037070004845015)(1.0092310946062784,1.0092310946062784)(0.9729963690576382,0.9729963690576382)(1.0704140508080533,1.0704140508080533)(0.9576468175292262,0.9576468175292262)(0.981807103459879,0.981807103459879)(0.9954171050511371,0.9954171050511371)(0.9907535797158871,0.9907535797158871)(0.9948068546283295,0.9948068546283295)
	};
	
    \nextgroupplot[title=Prostate]
	\addplot[only marks,mark=*,color=blue,fill opacity=0.0] coordinates
	{
	(0.9914040639777133,0.9914040639777133)(0.9386080597651408,0.9386080597651408)(0.9867304343018277,0.9867304343018277)(0.9763704471721462,0.9763704471721462)(0.9900702767145637,0.9900702767145637)(0.8936953706701201,0.8936953706701201)(1.0089242667243907,1.0089242667243907)(0.990659784658169,0.990659784658169)(0.9995385430493756,0.9995385430493756)(0.9748762301515423,0.9748762301515423)(0.9949405623925407,0.9949405623925407)(0.9539582472095905,0.9539582472095905)(0.9646206216579949,0.9646206216579949)(0.992545032544528,0.992545032544528)(1.0113927870831443,1.0113927870831443)(0.9944800486504187,0.9944800486504187)(0.9712674728710687,0.9712674728710687)(0.9555708797162842,0.9555708797162842)(0.9879624421947256,0.9879624421947256)
	};
	
    \nextgroupplot[title=Seminal vesicles]
	\addplot[only marks,mark=*,color=blue,fill opacity=0.0] coordinates
	{
	(0.8615908051235136,1.1962488563586464)(0.9705906514417156,0.9705906514417156)(0.9212003395120885,1.3247816444432279)(0.9067831798278725,0.9067831798278725)(0.8630021351606335,0.8630021351606335)(0.8547169102453318,0.8547169102453318)(0.8824753711669214,0.8824753711669214)(0.9358754990629841,0.9358754990629841)(0.890440386680988,1.255424274973147)(1.0281881700357351,1.0281881700357351)(0.9958004581318406,0.9958004581318406)(0.8715648942515022,1.1663302289305864)(0.9848041683105005,0.9848041683105005)(0.9785894728123583,0.9785894728123583)(0.9249046457825298,0.9249046457825298)(0.6688844254749352,1.211189220855302)(0.7317532766634562,1.1921611322808925)(0.9449646474297727,0.9449646474297727)(0.941607684529828,0.941607684529828)
	};
    
  \end{groupplot}
\end{tikzpicture}

  \caption{
Results of our experiments on the pelvic data in the Gold Atlas project~\cite{nyholm2018mr}.
For each ROI and patient, the associated $m$ is formed by finite sample approximation and used to compute $s^{\mathrm{A}_m}$ as defined by \eqref{eq:sa} and $s^{\mathrm{D}_m}$ as defined by \eqref{eq:sd}.
  }
  \label{fig:results_G}
\end{figure}

\input{res_article/results_L_pgf}

\begin{table}[htb!]
\caption{
Results of our experiments on the pelvic data in the Gold Atlas project (G)~\cite{nyholm2018mr} and the lung nodule data in LIDC-IDRI (L)~\cite{armato2011lung}.
For each ROI and patient, the associated $m$ is formed by finite sample approximation 
and used to compute $s^{\mathrm{A}_m}$ as defined by \eqref{eq:sa} and $s^{\mathrm{D}_m}$ as defined by \eqref{eq:sd}. 
}
\label{table:results}
\centering
\begin{tabular}{lllllllll}
\toprule
& 
\multicolumn{4}{c}{$\lVert s^{\mathrm{A}_m}\rVert_1/\lVert m\rVert_1$} &
\multicolumn{4}{c}{$\lVert s^{\mathrm{D}_m}\rVert_1/\lVert m\rVert_1$} \\
\cmidrule(r){2-5} 
\cmidrule(r){6-9} 
ROI & Mean & Std & Min & Max & Mean & Std & Min & Max\\
\midrule
(G) Urinary bladder
&  
1.004 & 0.009 & 0.991 & 1.035
& 
1.004 & 0.009 & 0.991 & 1.035 \\ 
(G) Rectum
&
0.994 & 0.047 & 0.912 & 1.094
&
0.994 & 0.047 & 0.912 & 1.094 \\
(G) Anal canal
& 
0.916 & 0.068 & 0.753 & 1.067 
&
1.075 & 0.182 & 0.877 & 1.560 \\
(G) Penile bulb
& 
0.929 & 0.072 & 0.696 & 1.022 
& 
1.065 & 0.157 & 0.863 & 1.365 \\
(G) Neurovascular b.
& 
0.778 & 0.110 & 0.461 & 0.928 
& 
1.267 & 0.115 & 1.070 & 1.481 \\
(G) Femoral head R
& 
0.988 & 0.011 & 0.963 & 1.006 
& 
0.988 & 0.011 & 0.963 & 1.006 \\
(G) Femoral head L
& 
0.994 & 0.022 & 0.958 & 1.070 
& 
0.994 & 0.022 & 0.958 & 1.070 \\
(G) Prostate
& 
0.978 & 0.027 & 0.894 & 1.011 
& 
0.978 & 0.027 & 0.894 & 1.011 \\
(G) Seminal vesicles
& 
0.903 & 0.085 & 0.669 & 1.028
&
1.029 & 0.142 & 0.855 & 1.325 \\
\midrule
(L) Lung nodules
& 
1.002 & 0.362 & 0.000 & 2.000
&
1.432 & 0.893 & 0.451 & 4.000 \\
\bottomrule
\end{tabular}
\end{table}


From our experiments we report the quantities $\lVert s^{\mathrm{A}_m}\rVert_1/\lVert m\rVert_1$ and $\lVert s^{\mathrm{D}_m}\rVert_1/\lVert m\rVert_1$, which in a relative sense describe how much the volume of the computed optimal segmentations with respect to Accuracy and Dice deviate from the expected target volume.
In Figure~\ref{fig:results_G} and Figure~\ref{fig:results_L}, these quantities are illustrated for each patient and ROI in scatter plots.
In Table~\ref{table:results}, aggregated statistics of these quantities with respect to all patients are shown.
By simple inspection it is clear that the volume of optimal segmentations to Accuracy and Dice often deviate significantly from the expected target volume.
For (G) the extreme cases are given by some marginal function $m$ for which $\lVert s^{\mathrm{A}_m}\rVert_1/\lVert m\rVert_1 \approx 0.5$ and some marginal function $m$ for which  $\lVert s^{\mathrm{D}_m}\rVert_1/\lVert m\rVert_1 \approx 1.5 $.
For (L) the extreme cases are given by some marginal function $m$ for which $\lVert s^{\mathrm{A}_m}\rVert_1/\lVert m\rVert_1 \approx 0$ and some marginal function $m$ for which  $\lVert s^{\mathrm{D}_m}\rVert_1/\lVert m\rVert_1 \approx 4$.


\section{Conclusion}
In this work, we have theoretically investigated the optimal segmentations with respect to the performance metrics Accuracy and Dice.
We have given a detailed rigorous characterization of the optimizers and upper and lower bounds on the volume of optimal segmentations.
Finally, we have shown the relevance of our theoretical observations in practice by comparing the volume of optimal segmentations with respect to the performance metrics to the expected volume, on two real world data sets.
We conclude that noise may cause optimal segmentations to have a volume that deviates significantly from the expected target volume and that the reason for this may be what metric is chosen, or implicitly, what metric a chosen training method targets.

\paragraph{Broader impacts:}
Formalizing the evaluation process of automated segmentation methods can be done in many ways, each with its pros and cons.
Even if this work can be interpreted as describing the problems with using Dice for this formalization, it can still paradoxically contribute to an unhealthy fixation of Dice as the gold standard for segmentation evaluation in medical image analysis.
This in turn can lead to that medical practitioners put too much faith in models that have been shown to perform well with respect to the metric on some test data.
One solution to this is to make sure that clinical practitioners using such models are educated in the problems associated with the metric.

\paragraph{Limitations:}
In order for the volume bounds to be sharp in Theorem~\ref{thm:accuracy} and Theorem~\ref{thm:dice}, we construct extreme cases of $m$.
These extreme cases might only be representable approximately with step functions for a particular choice of voxelization.
Consequently, the most extreme cases we can construct in a numerical setting may not be as extreme as those we have constructed in the continuous setting.
However, in medical image analysis it is common to deal with voxelizations of the order of $512\times 512\times 100$ voxels which means that the approximation error would be negligible.
Our work is also limited by the amount of experiments included.
Additional numerical experiments on a wider range of data sets would give a more comprehensive picture on the impact of different noise distributions.

\paragraph{Acknowledgement:}
The authors were supported by RaySearch Laboratories AB.

\bibliography{bibliography}

\end{document}


\maketitle

\tableofcontents

\clearpage
\section{Proofs}

\subsection{Generalities}
A lot of our work will be devoted to analyzing cumulative distribution functions and associated quantile functions.
There are several properties and identities that we will use that can be found in any standard textbook on the topic.
Here is an incomplete list.
Let $F(t)$, $t\in[0,1]$ be a cumulative distribution and let $F^{-1}(v) = \inf\{t : F(t) \geq v\}$ be the associated quantile function.
Then the following holds.
\begin{enumerate}
\item $F$ is non-decreasing and right-continuous
\item $F(0-) = 0$ and $F(1)=1$
\item $0\le F(t) \le 1$, $t\in[0,1]$.
\item $F^{-1}$ is non-decreasing and left-continuous.
\item $0\le F^{-1}(v)\le 1$, $v\in[0,1]$
\item $F^{-1}(F(t)) \le t$ and $F(F^{-1}(v)) \ge v$.
\item $F^{-1}(v)\le t$ if and only if $F(t) \le v$.
\item if $U$ is a uniform random variable over $[0,1]$, then $F^{-1}(U)$ has the cumulative distribution function $F$
\end{enumerate}

\subsection{Proof of Lemma 1}
\label{proof:lem:char}
\begin{proof}
Our proof consist of two parts.
In the first part, we show that the class of maximizers to 
\begin{align}
    \sup_{s\in\mathcal{S}_v} \int_\Omega s(\omega)m(\omega)\lambda(d\omega)
\end{align}
is given by $\mathcal{S}_{m,v}$.
In the second part, we show that if $s^*\in\mathcal{S}_{m,v}$, then 
\begin{align}
    \int_\Omega s^*(\omega)m(\omega) \lambda(d\omega) = \int_0^v (1-F_m^{-1}(u)) du.
\end{align}
This together shows the statement.

\paragraph{Part 1, optimality. }
Let $s^*\in\mathcal{S}_{m,v}$ and $s\in\mathcal{S}_v$.
Then $s^*$ and $s$ are feasible.
By construction, we have that there must exist some non-negative numbers $\epsilon_1,\epsilon_2\ge 0$ such that
\begin{align}
    \int_\Omega (s^*(\omega)-s(\omega))I\{m(\omega) < 1-F_m^{-1}(v)\} \lambda(d\omega) &=
    -\epsilon_1, \\
    \int_\Omega (s^*(\omega)-s(\omega))I\{m(\omega) = 1-F_m^{-1}(v)\} \lambda(d\omega) &=
    \epsilon_1-\epsilon_2, \\
    \int_\Omega (s^*(\omega)-s(\omega))I\{m(\omega) > 1-F_m^{-1}(v)\} \lambda(d\omega) &=
    \epsilon_2.
\end{align}
If $s\in\mathcal{S}_{m,v}$, then $\epsilon_1 = \epsilon_2 = 0$ and
consequently
\begin{align}
\int_\Omega (s^*(\omega)-s(\omega)) m(\omega) \lambda(d\omega) =0.
\end{align}
Otherwise, if $s\in \mathcal{S}_v\setminus \mathcal{S}_{m,v}$, then at least one of the inequaities $\epsilon_1>0$ and $\epsilon_2>0$ hold.
If $\epsilon_1 > 0$, it follows that
\begin{align}
\int_\Omega (s^*(\omega) - s(\omega)) I\{m(\omega) < 1-F_m^{-1}(v)\} m(\omega) \lambda(d\omega) &> -\epsilon_1 (1-F_m^{-1}(v))  \\
\int_\Omega (s^*(\omega) - s(\omega)) I\{m(\omega) = 1-F_m^{-1}(v)\} m(\omega) \lambda(d\omega) &= (\epsilon_1-\epsilon_2) (1-F_m^{-1}(v)) \\
\int_\Omega (s^*(\omega) - s(\omega)) I\{m(\omega) > 1-F_m^{-1}(v)\} m(\omega) \lambda(d\omega) &\ge \epsilon_2 (1-F_m^{-1}(v))
\end{align}
and similarly, if $\epsilon_2 > 0$, then
\begin{align}
\int_\Omega (s^*(\omega) - s(\omega)) I\{m(\omega) < 1-F_m^{-1}\} m(\omega) \lambda(d\omega) &\ge -\epsilon_1 (1-F_m^{-1}(v))  \\
\int_\Omega (s^*(\omega) - s(\omega)) I\{m(\omega) = 1-F_m^{-1}\} m(\omega) \lambda(d\omega) &= (\epsilon_1-\epsilon_2) (1-F_m^{-1}(v)) \\
\int_\Omega (s^*(\omega) - s(\omega)) I\{m(\omega) > 1-F_m^{-1}\} m(\omega) \lambda(d\omega) &> \epsilon_2 (1-F_m^{-1}(v)).
\end{align}
In either case, at least one inequality is strict and it follows that
\begin{align}
    \int_\Omega (s^*(\omega)-s(\omega)) m(\omega) \lambda(d\omega) > 0.
\end{align}
We conclude that $\mathcal{S}_{m,v}$ is the set of optimal segmentations where the supremum is attained.

\paragraph{Part 2, equality.}
Let $s^*\in \mathcal{S}_{m,v}$ and introduce
 \begin{align}
    C_1&\doteq \int_\Omega \bar m(\omega) I\{\bar m(\omega) \le F_m^{-1}(v)\}\lambda(d\omega).
\intertext{and}
    C_2&\doteq F_m^{-1}(v)(F_m(F_m^{-1}(v)) - v)
\end{align}

We now interpret $\bar m$ as a random variable on $(\Omega, \lambda)$ with cdf given by $F_m$ and then note by the quantile transform, that $F_m^{-1}(U)$ also has cdf given by $F_m$, where $U$ is uniformly distributed random variable on $[0,1]$.
It follows for $C_1$ that
\begin{align}
C_1 &= \mathbb{E}[F_m^{-1}(U)I\{F_m^{-1}(U) \le F_m^{-1}(v)\}] 
 = \mathbb{E}[F_m^{-1}(U)I\{U \le F_m(F_m^{-1}(v)) \}] 
 \intertext{and for $C_2$ that}
C_2&=
\mathbb{E}[F_m^{-1}(v) I\{ v < U\le F_m(F_m^{-1} (v)) \}]  =
\mathbb{E}[F_m^{-1}(U) I\{ v < U\le F_m(F_m^{-1} (v)) \}]
\end{align}
Together with the definition of $\mathcal{S}_{m,v}$ this yields
\begin{align}
    \int_\Omega s^*(\omega) \bar m(\omega) \lambda(d\omega) &= C_1-C_2 =\mathbb{E}[F_m^{-1}(U) I\{ U\le v \}] = \int_0^v F^{-1}(u) du 
\intertext{Consequently,}
\int_\Omega s^*(\omega) m(\omega) \lambda(d\omega)
    &=
    v-\left(\int_\Omega s^*(\omega) \lambda(d\omega) - \int_\Omega s(\omega) m(\omega) \lambda(d\omega)\right) \\
    &=
    v-\int_\Omega s^*(\omega) \bar{m}(\omega) \lambda(d\omega) \\
    &= v-\int_0^v F^{-1}(u) du \\
    &=\int_0^v(1-F^{-1}(u)) du 
\end{align}
This completes the proof.

\end{proof}

\subsection{Proof of Theorem 1}
\label{proof:thm:accuracy}
\begin{proof}
Consider the function
\begin{align}
    a_m(v) \doteq  v+1-\lVert m\rVert_1 - 2\int_0^v F_m^{-1}(u) du, \quad v\in[0,1].
\end{align} 
and note that by Lemma 1
\begin{align}
    \sup_{s\in\mathcal{S}} \mathrm{A}_m(s) &=
    \sup_{s\in\mathcal{S}} \int_\Omega [s(\omega) m(\omega) + \bar{s}(\omega) \bar{m}(\omega) ] \lambda(d\omega) \\
    &=
    \sup_{v\in[0,1]}\sup_{s\in\mathcal{S}_v} \int_\Omega [2s(\omega) m(\omega) + 1-s(\omega) -m(\omega) ] \lambda(d\omega) \\
    &=
      \sup_{v\in[0,1]}
   \left[1-\lVert m\rVert_1-v+2\sup_{s\in\mathcal{S}_v} \int_\Omega s(\omega) m(\omega) \lambda(d\omega)\right] \\
    &=
    \sup_{v\in[0,1]}
    \left[1-\lVert m\rVert_1-v+2 \int_0^v (1-F_m^{-1}(u)) du \right] \\
    &=
    \sup_{v\in[0,1]}
    \left[v+1-\lVert m\rVert_1-2 \int_0^v F_m^{-1}(u) du \right] \\
    &=
    \sup_{v\in[0,1]}
    a_m(v) \\
    &= 
    a_m(v^*), \quad v^*\in \mathcal{V}^*
\end{align}
where the supremum of $\mathrm{A}_m$ is attained for segmentations in $\cup_{v^* \in \mathcal{V}^*}\mathcal{S}_{m,v^*}$. 
Moreover, the function $a_m$ is differentiable a.e., with derivative given by
\begin{align*}
    \partial_v a_m(v) = 1-2F_m^{-1}(v).  
\end{align*}

\paragraph{Part 1, characterization.}
Since $F_m^{-1}$ is non-decreasing it follows that $\partial_v a_m$ is positive for $F_m^{-1}(v) < \frac{1}{2}$ and negative for $F_m^{-1}(v) > \frac{1}{2}$.
Therefore, the maximum of $a_m$ is attained for $v$ such that $F_m^{-1}(v) = \frac{1}{2}$. That is, the maximum is attained for $v \in (F_m(\frac{1}{2}-), F_m(\frac{1}{2})]$. For $v$ in this set, 
\begin{align}
    a_m(v) &= v+1-\lVert m\rVert_1 - 2\left( (v-F_m(\frac{1}{2}-))\frac{1}{2} + \int_0^{F_m(\frac{1}{2}-)} F_m^{-1}(u) du\right) \\
    &= F_m(\frac{1}{2}-) + 1-\lVert m\rVert_1 - \int_0^{F_m(\frac{1}{2}-)} F_m^{-1}(u) du. 
\end{align}

\paragraph{Part 2, bounds.}
We now prove the lower bound for $F_m(\frac{1}{2}-)$ and the upper bound for $F_m(\frac{1}{2})$.
For this, first note that if we interpret $\bar{m}$ as a random variable on $(\Omega,\lambda)$, with cdf given by $F_m$, and then note by the quantile transform, that $F_m^{-1}(U)$ also has cdf given by $F_m$ where $U$ is a uniformly distributed random variable on $[0,1]$, then
\begin{align}
    \int_0^1 (1-F_m^{-1}(u)) du = 1-\mathbb{E}[F^{-1}(U)] = 1-\int_\Omega \bar{m}(\omega) \lambda(d\omega) = \lVert m\rVert_1. 
\end{align}

Now, for the lower bound, note that, for $v \in [0,1]$,
\begin{align}
    1-\lVert m\rVert_1 = \int_0^1 F_m^{-1}(u) du \geq (1-v)F_m^{-1}(v),
\end{align}
which implies that
\begin{align}
    F_m^{-1}(v) \leq \frac{1-\lVert m\rVert_1}{1-v}.  
\end{align}
Consequently, for $\lVert m\rVert_1 \geq 1/2$, 
\begin{align}
    F_m^{-1}(v) \leq \frac{1-\lVert m\rVert_1}{1-v} < \frac{1}{2}, \quad 0 \leq v < 2\lVert m\rVert_1-1
\end{align}
and 
\begin{align}
F_m(\frac{1}{2}-) \geq \lim_{v \uparrow 1-2(1-\lVert m\rVert_1)} F_m(F_m^{-1}(v)) \geq 1-2(1-\lVert m\rVert_1) = 2\lVert m\rVert_1-1. 
\end{align}
For $\lVert m\rVert_1 < 1/2$, clearly $F_m(\frac{1}{2}-) \geq 0$.
The cases $\lVert m\rVert_1 \ge 1/2$ and $\lVert m\rVert_1<1/2$ together means that
\begin{align}
F(1/2-) \ge \max\{ 2\lVert m\rVert_1-1,0\}.
\end{align}

For the upper bound, note that, for $v \in [0,1]$,
\begin{align}
    1-\lVert m\rVert_1 = \int_0^1 F_m^{-1}(u) du \leq vF_m^{-1}(v) + 1-v,
\end{align}
which implies that
\begin{align}
    F_m^{-1}(v) \geq 1-\frac{\lVert m\rVert_1}{v}.  
\end{align}
Consequently, with $v = F_m(\frac{1}{2})$ 
\begin{align}
    1-\frac{\lVert m\rVert_1}{F_m(\frac{1}{2})} \leq  F_m^{-1}(F(\frac{1}{2})) \leq \frac{1}{2}, 
\end{align}
and it follows that $F_m(\frac{1}{2}) \leq 2 \lVert m\rVert_1$. For $\lVert m\rVert_1 \geq \frac{1}{2}$ we clearly have that $F_m(\frac{1}{2}) \leq 1$.
The cases $\lVert m\rVert_1 < 1/2$ and $\lVert m\rVert_1\ge 1/2$ together means that
\begin{align}
F(1/2) \le \min\{ 2\lVert m\rVert_1,1\}.
\end{align}

\paragraph{Part 3, sharpness.}
Recall that the domain $\Omega=[0,1]^n$  for some $n\ge 1$.
We use $(\omega_1,\dots,\omega_n)=\omega \in \Omega$ to denote the components.

To see that the lower bound is sharp, take $v \geq \frac{1}{2}$ and $m_0\in\mathcal{M}$ given by
\begin{align}
    m_0(\omega) = I_{[0,2v-1]}(\omega_1)+\frac{1}{2} I_{(2v-1,1]}(\omega_1), \quad \omega \in\Omega.
\end{align}
Note that
\begin{align}
    \lVert m_0\rVert_1 = (2v-1) + \frac{1}{2}(1-(2v-1)) = v.
\end{align}
Consequently,
\begin{align}
    F_{m_0}(t) = (2v-1)I_{[0,\frac{1}{2})}(t) + I_{[\frac{1}{2},1]}(t), \quad t \in [0,1],
\end{align}
which means that 
$F_{m_0}(\frac{1}{2}-) = 2v-1$.
Now instead take $v < 1/2$ and $m_0\in\mathcal{M}$ given by
\begin{align}
m_0(\omega) = v I_{[0,1]}(\omega_1), \quad \omega \in\Omega.
\end{align}
Note that 
\begin{align}
    \lVert m_0\rVert_1 = v(1-0) = v.
\end{align}
Furthermore,
\begin{align}
    F_{m_0}(t) = I_{[1-v,1]}(t), \quad t \in [0,1],
\end{align}
which means that
$F_{m_0}(\frac{1}{2}-) = 0$. 
Together, these cases say that for any $v\in [0,1]$, the $m_0$ formed by taking
\begin{align}
    m_0(\omega) = 
    \begin{cases}
    I_{[0,2v-1]}(\omega_1)+\frac{1}{2} I_{(2v-1,1]}(\omega_1), &\text{ if } v \ge 1/2,\\
    v I_{[0,1]}(\omega_1), &\text{ if } v < 1/2, 
    \end{cases} \quad \omega \in \Omega,
\end{align}
satisfies $\lVert m_0\rVert_1 = v$ and
\begin{align}
    F_{m_0}(\frac{1}{2}-) = \max\{2\lVert m_0\rVert_1 -1, 0\}.
\end{align}

To see that the upper bound is sharp, take $v < \frac{1}{2}$ and $m_1\in\mathcal{M}$ such that
\begin{align}
    m_1(\omega) = \frac{1}{2} I_{[0,2v]}(\omega_1), \quad \omega\in\Omega.
\end{align}
Note that
\begin{align}
    \lVert m_1\rVert_1 = \frac{1}{2}(2v-0) = v.
\end{align}
Furthermore,
\begin{align}
    F_{m_1}(t) = 2v I_{[\frac{1}{2},1)}(t) + I_{\{1\}}(t), \quad t \in [0,1],
\end{align}
which means that
$F_{m_1}(\frac{1}{2}) = 2v$.
Now instead take $v\geq 1/2$ and $m_1\in\mathcal{M}$ given by
\begin{align}
    m_1(\omega) = v, \quad \omega \in \Omega.
\end{align}
Note that 
\begin{align}
    \lVert m_1\rVert_1 = v.
\end{align}
Furthermore
\begin{align}
    F_{m_1}(t) = I_{[1-v,1]}(t), \quad t \in [0,1],
\end{align}
which means that $F_{m_1}(\frac{1}{2}) = 1$. 
Together, these cases say that for any $v\in[0,1]$, the $m_1$ formed by taking
\begin{align}
    m_1(\omega) = 
    \begin{cases}
    \frac{1}{2} I_{[0,2v]}(\omega_1), \quad &\text{ if } v< 1/2 \\
     v, \quad &\text{ if } v \ge 1/2,
    \end{cases} \quad \omega \in \Omega,
\end{align}
satisfies $\lVert m_1\rVert_1=v$ and
\begin{align}
    F_{m_1}\left(\frac{1}{2}\right) = \min\{2\lVert m_1\rVert_1, 1\}.
\end{align}
This completes the proof. 
\end{proof}

\subsection{Proof of Theorem 2}
\label{proof:thm:dice}
\begin{proof}
Consider the function
\begin{align}
    d_m(v) \doteq \frac{2\int_0^v 1-F_m^{-1}(u) du}{\lVert m\rVert_1+v}, \quad v\in[0,1].
\end{align}
and note that by Lemma 1 
\begin{align}
    \sup_{s \in \mathcal{S}} D(s) &= \sup_{v \in [0,1]} \frac{\sup_{s \in \mathcal{S}_v} 2 \int_\Omega s(\omega)m(\omega)\lambda(d\omega) }{\lVert m\rVert_1 + v} \\
    &= \sup_{v \in [0,1]} \frac{2 \int_0^v 1-F_m^{-1}(u) du }{\lVert m\rVert_1 + v} \\
    &= \sup_{v \in [0,1]} d_m(v) \\
    &= d_m(v^*),\quad v^* \in \mathcal{V}^*,
\end{align}
where the supremum of $\mathrm{D}_m$ is attained for segmentations in $\cup_{v^* \in \mathcal{V}^*}\mathcal{S}_{m,v^*}$. 
Note that $d_m$ is continuous
and hence attains its maximum on the compact
set $[0,1]$. By continuity, the set $\mathcal{V}^*\subset [0,1]$ where the maximum is attained is closed and therefore compact.
Moreover, $d_m$ is differentiable a.e.\ with derivative given by
\begin{align}
 \partial_v d_m(v) = \frac{2(1-F_m^{-1}(v)) - d_m(v)}{\lVert m\rVert_1 + v}. 
\end{align}
Consider the function $\delta$ given by
\begin{equation}\label{eq:delta}
 \delta(v) = \lVert m\rVert_1 +\int_0^v F_m^{-1}(u) du  - (\lVert m\rVert_1 + v)F_m^{-1}(v), \quad v \in [0,1]. 
\end{equation}
The interest in the function $\delta$ comes from the fact that it has the same sign as $\partial_v d_m$ but is somewhat easier to work with. Indeed, note that
$\delta(v) = \frac{(\lVert m\rVert_1 + v)^2}{2}\partial_v d_m(v)$ and hence,  $\partial_v d_m(v) > 0$ if and only if $\delta(v) > 0$, and similarly, $\partial_v d_m(v) < 0$ if and only if $\delta(v) < 0$. 
\paragraph{Part 1, characterization.}
Note first that $\delta$ is left-continuous and non-increasing on $[0,1]$.
That $\delta$ is left-continuous follows by construction since $F_m^{-1}$ is left continuous.
That $\delta$ is non-increasing on $[0,1]$ follows since $F_m^{-1}$ is non-decreasing and consequently that for $0 \leq v_0 < v_1 \leq 1$ 
\begin{align}\label{eq:deltadiff}
    \delta(v_1) - \delta(v_0)
    &= \lVert m\rVert_1(F_m^{-1}(v_0) - F_m^{-1}(v_1)) + (v_0F_m^{-1}(v_0) - v_1 F_m^{-1}(v_1)) - \int_{v_0}^{v_1} F_m^{-1}(u) du  \\
    &\le \lVert m\rVert_1\cdot 0 + 0 + 0 \\
    &\le 0.
\end{align}
Using these properties, it follows that there are three possibilities.
\begin{enumerate}[(i)]
    \item $\delta(v) > 0$ for all $v \in [0,1)$, in which case $d_m$ is strictly increasing on $[0,1)$ and attains its maximum at $v^* = 1$. In this case $\mathcal{V}^*= \{1\}$. 
    \item There is a half-open interval $(a,b] \subset [0,1]$ where $\delta = 0$. By continuity $d_m$ is constant on the closed interval $\mathcal{V}^* = [a,b]$ and attains its maximum on $\mathcal{V}^*$.  
    \item There is a unique point $v^* \in [0,1]$ such that $\delta(v) > 0$ for $v < v^*$ and $\delta(v) < 0$, for $v > v^*$. In this case $\mathcal{V}^*= \{v^*\}$ and $d$ attains its maximum at $v^*$.
\end{enumerate}
In each case $\mathcal{V}^*$ is a closed subinterval of $[0,1]$.
Now we prove that it is exactly the interval  $\mathcal{V}^* = [F_m((1-d_m^*/2)-), F_m(1-d_m^*/2)]$ where $d_m^*=d_m(v^*)$ for $v^*\in\mathcal{V}^*$ and $F_m((1-d_m^*/2)-) = \lim_{t \uparrow 1-d_m^*/2} F_m(t)$ denotes the left limit.
To this end, consider the three cases (i), (ii), and (iii) separately. 

In case (i) $\mathcal{V}^* = \{1\}$, so it is sufficient to show that $F_m((1-\frac{d_m^*}{2})-) = 1$.  Recall that $\delta(v) > 0$ for $v < 1$ and $\delta$ is left-continuous it follows that $\delta(1) \geq 0$ and by definition of $\delta$ that
\begin{align}
    F_m^{-1}(1) \leq 1-\frac{d_m(1)}{2}=1-\frac{d_m^*}{2}.
\end{align}
There are two cases to consider. If $F_m^{-1}(1) = 1-\frac{d_m^*}{2}$, then $\delta(1) = 0$ and we claim that
$F_m^{-1}(v) < F_m^{-1}(1)$, for $v < 1$.
Indeed, otherwise $F_m^{-1}(v_0) = F_m^{-1}(1)$ for some $v_0 < 1$, which implies that $F_m^{-1}(v) = F_m^{-1}(1)$ for all $v \in [v_0,1]$ and, by \eqref{eq:deltadiff}, $\delta(v)$ is constant on $[v_0,1]$.
But since by assumption $\delta(v)>1$ for $v\in[v_0,1)$, this means that $\delta(1)>0$ which is a contradiction.
In the other case, if $F_m^{-1}(1) < 1-\frac{d_m^*}{2}$, then 
$F_m^{-1}(v) < 1-\frac{d_m^*}{2}$, since $F_m^{-1}$ is non-decreasing. In both cases it holds that $F_m^{-1}(v) < 1-\frac{d_m^*}{2}$, for $v < 1$, and consequently, 
\begin{align}
    1 = \lim_{v \uparrow 1} v \leq \lim_{v \uparrow 1}  F_m(F_m^{-1}(v)) \leq F_m((1-\frac{d_m^*}{2})-). 
\end{align}
This completes the proof in case (i). 

In case (ii), with $\mathcal{V}^* = [a,b]$, $a < b$, we show first that $a = F_m((1-\frac{d_m^*}{2})-)$ by showing the inequalities $a \leq F_m((1-\frac{d_m^*}{2})-)$ and $a \geq F_m((1-\frac{d_m^*}{2})-)$. The first inequality is similar to the proof of case (i) above. For $v < a$, $\delta(v) > 0$, and left-continuity of $\delta$ implies that $\delta(a) \geq 0$ and it follows that
\begin{align}
    F_m^{-1}(a) \leq 1-\frac{d_m(a)}{2} = 1-\frac{d_m^*}{2}.
\end{align}
There are two cases to consider. If $F_m^{-1}(a) = 1-\frac{d_m^*}{2}$, then $\delta(a) = 0$ and we claim that
$F_m^{-1}(v) < F_m^{-1}(a)$, for $v < a$. Indeed, otherwise $F_m^{-1}(v_0) = F_m^{-1}(a)$ for some $v_0 < a$, which implies that $F_m^{-1}(v) = F_m^{-1}(a)$ for all $v \in [v_0,a]$ and, by \eqref{eq:deltadiff}, $\delta(v)$ is constant on $[v_0,a]$.
But since by assumption $\delta(v)>0$ for $v\in[v_0,a)$, this means that $\delta(a) > 0$ which is a contradiction.
In the other case, if $F_m^{-1}(a) < 1-\frac{d_m^*}{2}$, then 
$F_m^{-1}(v) < 1-\frac{d_m^*}{2}$, since $F_m^{-1}$ is non-decreasing. In both cases it holds that $F_m^{-1}(v) < 1-\frac{d_m^*}{2}$, for $v < a$, and consequently, 
\begin{align}
    a = \lim_{v \uparrow a} v \leq \lim_{v \uparrow a}  F_m(F_m^{-1}(v)) \leq F_m((1-\frac{d_m^*}{2})-). 
\end{align}

To show the reverse inequality  $a \geq F_m((1-\frac{d_m^*}{2})-)$, note that for $t < 1-\frac{d_m^*}{2}$ and $v < F_m(t)$, it holds that
\begin{align}
    F_m^{-1}(v) \leq t < 1- \frac{d_m^*}{2} \leq 1-\frac{d_m(v)}{2},
\end{align}
which implies that $\delta(v) > 0$. Consequently, $v \leq a$ and we conclude that $F_m((1-\frac{d_m^*}{2})-) \leq a$. 

To complete the proof in case (ii) it remains to show that
$F_m((1-\frac{d_m^*}{2})) = b$. For $v \in (a,b]$ it holds that $\delta(v) = 0$ and it follows that $F_m^{-1}(v) = 1-\frac{d_m^*}{2}$. Therefore, 
\begin{align}
    F_m(1-\frac{d_m^*}{2}) = F_m(F_m^{-1}(v)) \geq v.
\end{align}
By taking the limit as $v \uparrow b$ we conclude that $b \leq F_m(1-\frac{d_m^*}{2})$. For the reverse inequality, since
\begin{align}
    F_m^{-1}(F_m(1-\frac{d_m^*}{2})) \leq 1-\frac{d_m^*}{2}, 
\end{align}
it follows that $\delta(F_m(1-\frac{d_m^*}{2})) \geq 0$ and we conclude that $F_m(1-\frac{d_m^*}{2}) \leq b$. 

In case (iii), with $\mathcal{V}^* = \{v^*\}$, it is sufficient to prove that 
\begin{align}
v^* \leq F_m((1-\frac{d_m^*}{2})-) \leq  F_m(1-\frac{d_m^*}{2}) \leq v^*.     
\end{align}
The proof of $v^* \leq F_m((1-\frac{d_m^*}{2})-)$ is identical to the proof of $a \leq F_m((1-\frac{d_m^*}{2})-)$ in (ii) and is therefore omitted. To prove $v^* \geq F_m((1-\frac{d_m^*}{2}))$, recall that $\delta(v) > 0$ for $v < v^*$, $\delta(v^*) \geq 0$, by left-continuity of $\delta$, and $\delta(v) < 0$, $v > v^*$. Since
\begin{align}
    F_m^{-1}(F_m(1-\frac{d_m^*}{2})) \leq 1-\frac{d_m^*}{2}, 
\end{align}
it follows that $\delta(F_m(1-\frac{d_m^*}{2})) \geq 0$ and we conclude that $F_m(1-\frac{d_m^*}{2}) \leq v^*$.

This completes the proof of $\mathcal{V}^* = [F_m((1-d_m^*/2)-), F_m(1-d_m^*/2)]$.
\paragraph{Part 2, bounds.}
We now show that $\mathcal{V}^*\in [\lVert m\rVert_1^2, 1]$.
Note that the upper bound is the upper bound of the feasibility set $v\in[0,1]$ so it is already done.
For the lower bound, note that
\begin{align}
    1-\lVert m\rVert_1 = \int_0^1 F_m^{-1}(u) du \geq \int_{1-v}^1 F_m^{-1}(u) du \geq (1-v)F_m^{-1}(v), 
\end{align}
which implies that
\begin{align}
    F_m^{-1}(v) \leq \frac{1-\lVert m\rVert_1}{1-v}, \quad v \in [0,1]. 
\end{align}
Therefore, 
\begin{align}
    \delta(v) &\geq \lVert m\rVert_1(1-F_m^{-1}(v)) - v F_m^{-1}(v) 
 \geq \lVert m\rVert_1 - (\lVert m\rVert_1 + v)\frac{1-\lVert m\rVert_1}{1-v} = \frac{\lVert m\rVert_1^2 - v}{1-v}. 
\end{align}
We conclude that $\delta(v) > 0$ for $v < \lVert m\rVert_1^2$, which completes the proof.

\paragraph{Part 3, sharpness.}
Recall that the domain $\Omega=[0,1]^n$  for some $n\ge 1$.
We use $(\omega_1,\dots,\omega_n)=\omega \in \Omega$ to denote the components.

To prove that the bounds are sharp, it is enough to consider one example.
Take $v'\in (0,1]$ and $m\in\mathcal{M}$ such that
\begin{align}
m(\omega) =
I_{[0,v'^2]}(\omega_1) + \frac{v'}{1+v'}I_{(v'^2,1]}(\omega_1)
,\quad \omega\in\Omega.
\end{align}
Note that
\begin{align}
    \lVert m\rVert_1 = (v'^2-0) + \frac{v'}{1+v'}(1-v'^2) = v'.
\end{align}
Furthermore,
\begin{align}
    F_{m}(t) = v'^2 I_{[0,\frac{1}{1+v'})}(t) + I_{[\frac{1}{1+v'},1]}(t), \quad t \in [0,1],
\end{align}
or equivelently,
\begin{align}
    F_{m}(t) = \lVert m\rVert_1^2 I_{[0,\frac{1}{1+\lVert m\rVert_1})}(t) + I_{[\frac{1}{1+\lVert m\rVert_1},1]}(t), \quad t \in [0,1],
\end{align}
which means that
\begin{align}
    F_m^{-1}(v) = \frac{1}{1+\lVert m\rVert_1}I_{(\lVert m\rVert_1^2,1]}(v), \quad v \in [0,1].
\end{align}
Then for $v\in [0,1]$
\begin{align}
 \delta(v) &= \lVert m\rVert_1 +\int_0^v F_m^{-1}(u) du  - (\lVert m\rVert_1 + v)F_m^{-1}(v) \\
 &=  \lVert m\rVert_1 + \int_0^v \frac{1}{1+\lVert m\rVert_1} I_{(\lVert m\rVert_1^2,1]}(u) du - (\lVert m\rVert_1+v)\frac{1}{1+m} I_{(\lVert m\rVert_1^2,1]}(v) \\
 &=
 \lVert m\rVert_1 + \frac{1}{1+\lVert m\rVert_1} (v-\lVert m\rVert_1^2) I_{(\lVert m\rVert_1^2,1]}(v) - \frac{\lVert m\rVert_1+v}{1+\lVert m\rVert_1}I_{(\lVert m\rVert_1^2,1]}(v) \\
 &=
 \lVert m\rVert_1 - \lVert m\rVert_1I_{(\lVert m\lVert_1^2,1]} \\
 &=
 \lVert m\rVert_1 I_{[0,\lVert m\rVert_1^2]} 
\end{align}
Hence, $\delta(v) = 0$ for $v > \lVert m\rVert_1^2$ and $d$ attains its maximum on $\mathcal{V}^* = [\lVert m\rVert_1^2,1]$.
Since we we can choose any $v'\in(0,1]$ so that $\lVert m\rVert_1=v'$, this completes the proof.

\end{proof}

\subsection{Proof of Theorem 3}
\label{proof:thm:relation}
\begin{proof}
If we interpret $\bar{m}$ as a random variable on $(\Omega,\lambda)$, with cdf given by $F_m$, and then note by the quantile transform, that $F_m^{-1}(U)$ also has cdf given by $F_m$ where $U$ is a uniformly distributed random variable on $[0,1]$, it follows that
\begin{align}
    \int_0^1 (1-F_m^{-1}(u)) du = 1-\mathbb{E}[F^{-1}(U)] = 1-\int_\Omega \bar{m}(\omega) \lambda(d\omega) = \lVert m\rVert_1. 
\end{align}
Now, as in Theorem 2, consider the function
\begin{align}
    d_m(v) = \frac{2\int_0^v (1-F_m^{-1}(u)) du}{\lVert m\rVert_1+v}, \quad v\in[0,1].
\end{align}
and note that by Lemma 1 
\begin{align}
    \sup_{s \in \mathcal{S}} \mathrm{D}_m(s) &= \sup_{v \in [0,1]} \frac{\sup_{s \in \mathcal{S}_v} 2 \int_\Omega s(\omega)m(\omega)\lambda(d\omega) }{\lVert m\rVert_1 + v} \\
    &= \sup_{v \in [0,1]} \frac{2 \int_0^v (1-F_m^{-1}(u)) du }{\lVert m\rVert_1 + v} \\
    & = \sup_{v \in [0,1]} d_m(v) \\
    &\doteq d_m^*.
\end{align}
This means
\begin{align}
    d_m(v) = \frac{2\int_0^v (1-F_m^{-1}(u) dv)}{\lVert m\rVert_1 + v} 
    \le \frac{\int_0^1 (1-F_m^{-1}(u) dv) + \int_0^v 1 dv}{\lVert m\rVert_1 + v} 
    = 1, \quad v \in [0,1],
\end{align}
and consequently that $d_m^*\le 1$.
We now show that
\begin{align}
\sup \mathcal{V}^{\mathrm{A}_m} = F_m(1/2)\le F_m((1-d_m^*/2)-) = \inf \mathcal{V}^{\mathrm{D}_m}
\end{align}
separately for the cases $d_m^*<1$ and $d_m^*=1$.
First, assume that $d_m^*<1$.
Then there exist some $\epsilon>0$ such that $d_m^*<1-2\epsilon$, hence
\begin{align}
    1-d_m^*/2 > 1-(1-2\epsilon)/2 = 1/2+\epsilon,
\end{align}
and since $F_m$ is non-decreasing
\begin{align}
    F_m((1-d_m^*/2)-) \ge F_m((1/2+\epsilon)-) \ge F_m(1/2).
\end{align}
Secondly, assume that $d_m^* = 1$.
Then there must exist some $v$ such that $d_m(v)=1$, or equivalently
\begin{align}
    \int_0^v (1-F_m^{-1}(u)) du = \frac{1}{2}(\lVert m\rVert_1+v).
\end{align}
If $v<\lVert m\rVert_1$, then
\begin{align}
    v=\int_0^v 1 dv \ge \int_0^v(1-F_m^{-1}(u))du = \frac{1}{2}(\lVert m\rVert_1+v)> v
\end{align}
which is a contradiction and if $v>|m|$ then
\begin{align}
    \lVert m\rVert_1=\int_0^1 (1-F_m^{-1}(u)) du \ge \int_0^v(1-F_m^{-1}(u))du = \frac{1}{2}(\lVert m\rVert_1+v) > \lVert m\rVert_1
\end{align}
which also is a contradiction.
Consequently, we must have that $v=\lVert m\rVert_1$.
This in turn means that $F_m^{-1}$ needs to satisfy
\begin{align}
    \int_0^{\lVert m\rVert_1} (1-F_m^{-1}(u)) du = \lVert m\lVert_1.
\end{align}
which can only be the case if
\begin{align}
    F_m^{-1}(v) = I_{(\lVert m\rVert_1,1]}(v), \quad v\in[0,1].
\end{align}
This means that
\begin{align}
    F_m(t) = \lVert m\rVert_1I_{[0,1)}(t) + I_{\{1\}}(t).
\end{align}
and finally that
\begin{align}
    F_m((1-d_m^*/2)-) = F_m(1/2) = \lVert m\rVert_1.
\end{align}
This completes the proof.
\end{proof}

\subsection{Proof of Theorem 4}
\label{proof:thm:constrained}
\begin{proof}
By Lemma 1, the maximizers to
\begin{align}
    \sup_{s\in\mathcal{S}_v} \mathrm{A}_m(s)
    &=
    \sup_{s\in\mathcal{S}_v} \int_\Omega [s(\omega) m(\omega) + \bar{s}(\omega) \bar{m}(\omega) ] \lambda(d\omega) \\
    &=
   1-\lVert m\rVert_1-v+2\sup_{s\in\mathcal{S}_v} \int_\Omega s(\omega) m(\omega) \lambda(d\omega)
\intertext{are given by $\mathcal{S}_{m,v}$, and the maximizers to }
    \sup_{s \in \mathcal{S}_v} \mathrm{D}_m(s) &=
    \frac{2\int_\Omega s(\omega)m(\omega) \lambda(d\omega)}{\lVert m\rVert_1 + \lVert s\rVert_1} \\
    &= \frac{ 2}{\lVert m\rVert_1 + v} \sup_{s \in \mathcal{S}_v} \int_\Omega s(\omega)m(\omega)\lambda(d\omega)
\end{align}
are given by $\mathcal{S}_{m,v}$.
This completes the proof.
\end{proof}

\subsection{Proof of Theorem 5}
\label{proof:thm:threshold}
\begin{proof}
First note that $\lVert s_0\rVert_1 = \int_\Omega I\{\bar m(\omega) < t\}\lambda(d\omega) = F_m(t-)$ and 
$\lVert s_1\rVert_1 = \int_\Omega I\{\bar m(\omega) \leq  t\}\lambda(d\omega) = F_m(t)$.
We now prove the if and only if cases separately.
\paragraph{Part 1, $\Rightarrow$.}
Take any $s\in \cup_{v\in [F(t-),F(t)]}\mathcal{S}_{m,v}$.
To prove the upper bound $s(\omega)\le s_1(\omega), \lambda$-a.e., let $A = \{\omega : s(\omega) > s_1(\omega)\}$.
Since both $s$ and $s_1$ are binary it follows that 
\begin{align}
    I\{\omega \in A\} = s(\omega)\bar s_1(\omega), \quad \omega \in \Omega.
\end{align}
By definition of $\mathcal{S}_{m,v}$ and, since $v \le F_m(t)$ implies that $F_m^{-1}(v) \le F_m^{-1}(F_m(t))\leq t$, it follows that
\begin{align}
    0 = \int_\Omega s(\omega) I\{\bar m(\omega) > F_m^{-1}(v)\} \lambda(d\omega)
    \geq \int_\Omega s(\omega) I\{\bar m(\omega) > t \} \lambda(d\omega) \geq 0.
\end{align}
That is, 
\begin{align}
    \int_\Omega s(\omega) I\{\bar m(\omega) > t \} \lambda(d\omega) = 0.
\end{align}
Consequently, 
\begin{align}
    \lambda(A) = \int_\Omega s(\omega) \bar s_1(\omega) \lambda(d\omega) = \int_\Omega s(\omega) I\{\bar m(\omega) > t\} \lambda(d\omega) = 0.
\end{align} 
To prove the lower bound $s_0(\omega)\le s(\omega), \lambda$-a.e., let $B = \{\omega : s_0(\omega) > s(\omega)\}$. Since $s_0$ and $s$ are binary, it follows that
\begin{align}
    I\{\omega \in B\} = s_0(\omega)\bar s(\omega) = \bar s(\omega) I \{\bar m(\omega) < t\}, \qquad \omega \in \Omega.
\end{align}
Therefore, it is sufficient to show that
\begin{align}\label{eq:show0}
    \int_\Omega \bar s(\omega) I\{\bar m(\omega) < t\} \lambda(d\omega) = 0. 
\end{align}
We know from the definition of $\mathcal{S}_{m,v}$ that 
\begin{align}
    \int_\Omega \bar s(\omega) I\{\bar m(\omega) < F_m^{-1}(v)\} \lambda(d\omega) = 0. 
\end{align}
Consequently, with $v \in [F_m(t-),F_m(t)]$ it follows that $F_m^{-1}(v) \leq t$ and
\begin{align}
    0 &\leq  \int_\Omega \bar s(\omega) I\{\bar m(\omega) < t\} \lambda(d\omega) \\
    &= \underbrace{\int_\Omega \bar s(\omega) I\{\bar m(\omega) < F_m^{-1}(v)\} \lambda(d\omega)}_{= 0} 
    + \int_\Omega \bar s(\omega) I\{F_m^{-1}(v) \leq \bar m(\omega) < t\} \lambda(d\omega) \\
    & \leq \int_\Omega I\{F_m^{-1}(v) \leq \bar m(\omega) < t\} \lambda(d\omega) \\
    &= F_m(t-) - F_m(F_m^{-1}(v)) \\
    & \leq F_m(t-) - v \\
    & \leq 0.
\end{align}
 We conclude that \eqref{eq:show0} holds.
 
\paragraph{Part 2, $\Leftarrow$.}
Take any $s\in\mathcal{S}$ such that $s_0(\omega) \leq s(\omega) \leq s_1(\omega) \lambda$-a.e. Since, 
\begin{align}
    F_m(t-) = \|s_0\|_1 \leq \|s\|_1 \leq \|s_1\|_1 = F(t), 
\end{align}
it follows that $s \in \mathcal{S}_v$, with $v \in [F_m(t-), F_m(t)]$. Furthermore, with $v =\|s\|_1$,
\begin{align}
    0 &\leq \int_\Omega s(\omega) I\{m(\omega) < 1-F_m^{-1}(v) \} \lambda(d\omega) \\
    &= \int_\Omega s(\omega) I\{ \bar m(\omega) > F_m^{-1}(v) \} \lambda(d\omega) \\
    & \leq \int_\Omega s(\omega) I\{ \bar m(\omega) > F_m^{-1}(F_m(t-)) \} \lambda(d\omega) \\
    & \leq  \int_\Omega s_1(\omega) I\{\bar m(\omega) > F_m^{-1}(F_m(t-)) \} \lambda(d\omega) \\
    & = \int_\Omega  I\{F_m^{-1}(F_m(t-)) < \bar m(\omega) \leq  t\} \lambda(d\omega) \\
    &= F_m(t) - F_m(F_m^{-1}(F_m(t-))) = 0,
\end{align}
where the last equality follows since $F_m$ is a cumulative distribution function. Similarly, 
\begin{align}
    0 &\leq \int_\Omega \bar s(\omega) I\{m(\omega) > 1-F_m^{-1}(v) \} \lambda(d\omega) \\
    &= \int_\Omega (1-s(\omega)) I\{ \bar m(\omega) < F_m^{-1}(v) \} \lambda(d\omega) \\
    & \leq \int_\Omega (1-s(\omega)) I\{ \bar m(\omega) < F_m^{-1}(F_m(t)) \} \lambda(d\omega) \\
    & \leq \int_\Omega (1-s_0(\omega)) I\{\bar m(\omega) < F_m^{-1}(F_m(t)) \} \lambda(d\omega) \\
    & = \int_\Omega  I\{t \leq \bar m(\omega) < F_m^{-1}(F_m(t)) \} \lambda(d\omega) \\
    &= F_m(F_m^{-1}(F_m(t))-) -F_m(t-) \leq 0.
\end{align}
We conclude that any segmentation $s\in\mathcal{S}$ satisfying $s_0(\omega) \leq s(\omega) \leq s_1(\omega) \lambda$-a.e.
belongs to $\mathcal{S}_{m,v}$ with $v \in [F_m(t-),F_m(t)]$.

This completes the proof.
\end{proof}

\section{Experiments}
This section contains information about the conducted experiments.
However, the code and details of how to reproduce the results are found in the GitHub repository:
\begin{center}
    \url{https://github.com/marcus-nordstrom/optimal-solutions-to-accuracy-and-dice}
\end{center}

\label{experimental:details}

\subsection{Experiment G}

\subsubsection{Details on experiment}
\begin{figure}[htb!]
    \centering

\begin{tikzpicture}
  	\begin{axis}[
  	    name=plot11,height=6cm,width=6cm,
		xlabel=$\omega_1$, xtick={0,0.25,0.5,0.75,1.0},xmin=0,xmax=1 ,
		ylabel=$\omega_2$, ytick={0,0.25,0.5,0.75,1.0}, ymin=0,ymax=1, axis on top]
         \addplot graphics[xmin=0,xmax=1,ymin=0,ymax=1] {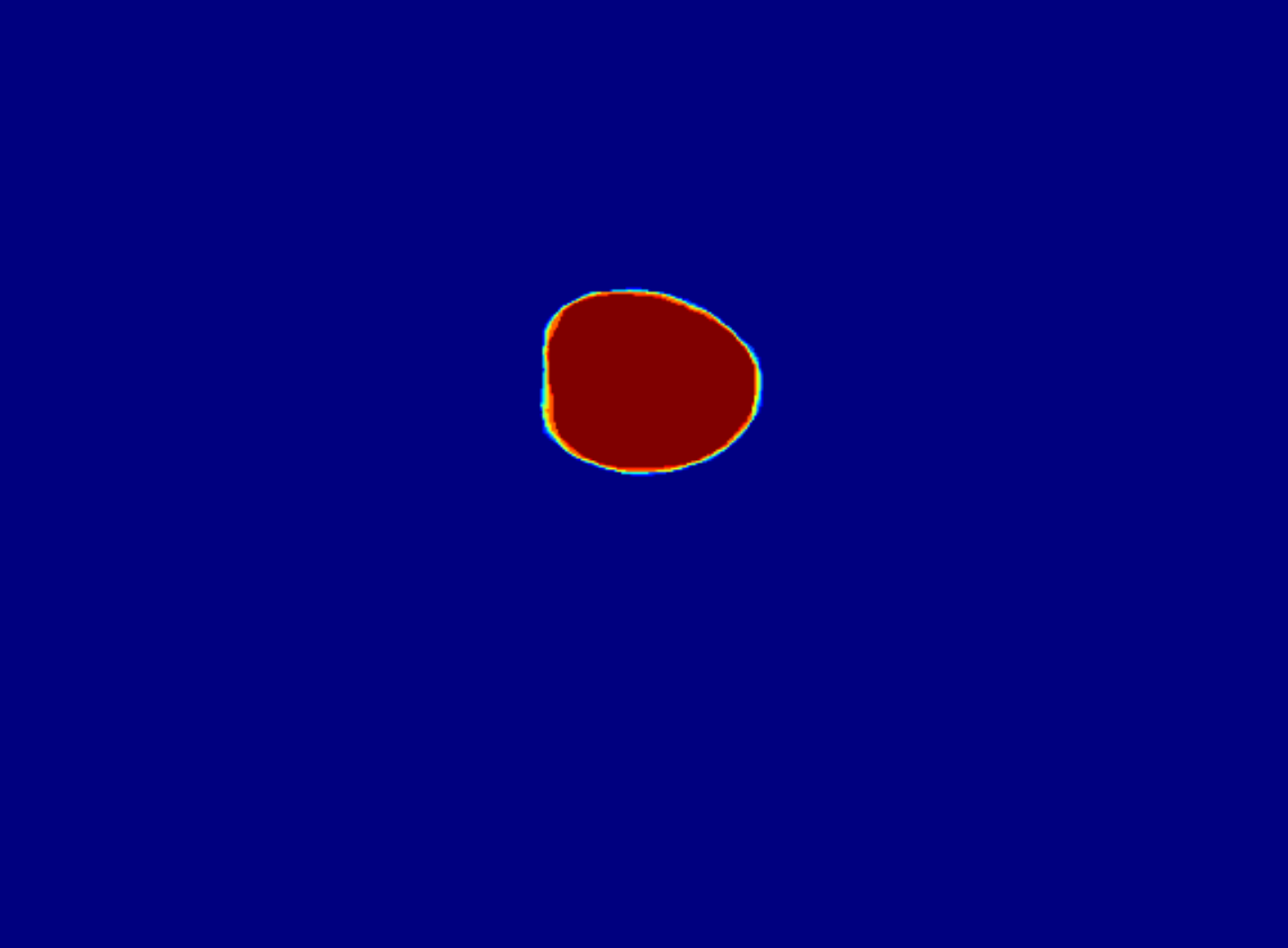};
	\end{axis}
  \begin{axis}[name=plot12,at={($(plot11.east)+(1.5cm,0)$)},anchor=west,height=6cm,width=6cm,
		xlabel=$\omega_1$, xtick={0,0.25,0.5,0.75,1.0},xmin=0,xmax=1 ,
		ylabel=$\omega_2$, ytick={0,0.25,0.5,0.75,1.0}, ymin=0,ymax=1, axis on top]
         \addplot graphics[xmin=0,xmax=1,ymin=0,ymax=1] {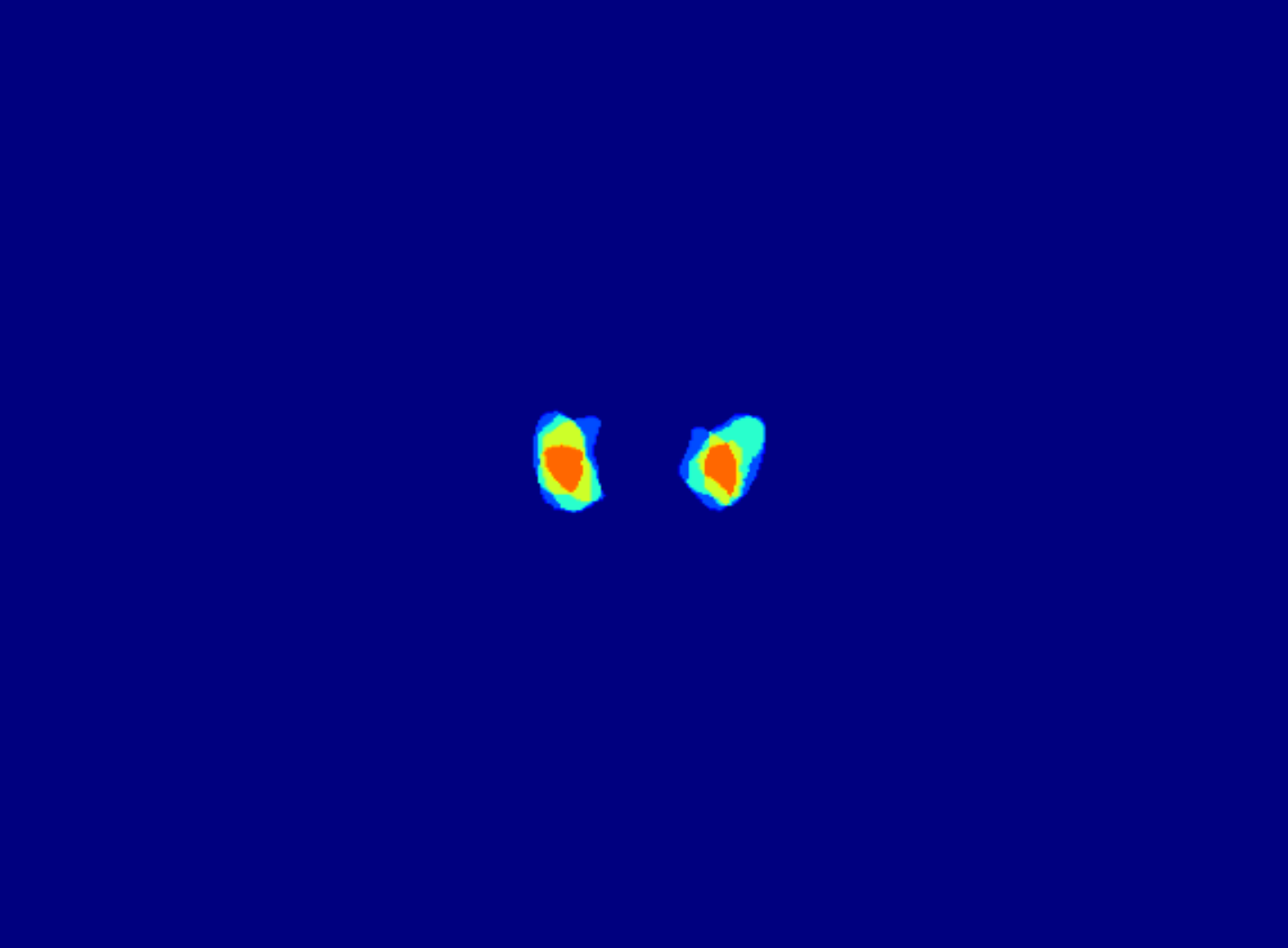};
  \end{axis}
  
  \begin{axis}[name=plot12,at={($(plot12.east)+(0.75cm,0)$)},anchor=west,height=6cm,width=2cm,
		xtick=\empty,
		ytick={0,0.25,0.5,0.75,1.0},
		yticklabel pos = right, xmin=0, xmax=1,ymin=0,ymax=1,axis on top]
 \addplot graphics[xmin=0,xmax=1,ymin=0,ymax=1] {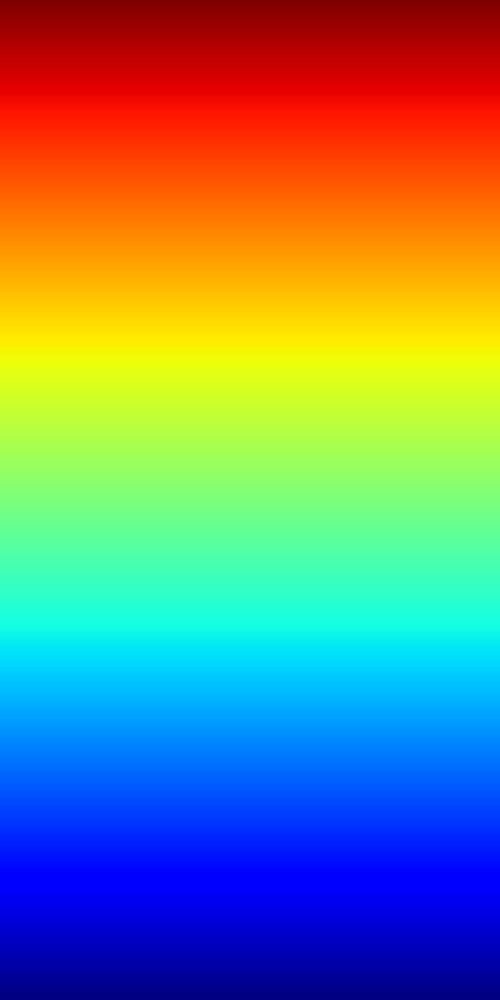};
  \end{axis}
\end{tikzpicture}

    \caption{Example of a slice from two marginal functions. The number on the axis are indices associated with the pixels. To the left, an Urinary bladder and to the right Neurovascular bundles. Note that there is almost no disagreement in the first case where as there is lots of disagreement in the second case.}%
    \label{fig:ill}%
\end{figure}

The data consist of 19 patients with 9 ROIs (region of interest) where each ROI has been delineate by 5 separate practitioners.
This leads to a total of 855 segmentations.
Let $1 \le r \le 9$ be an index of the ROIs, and $1 \le p \le 19$ be an index of the patients.
We then think of each ROI for each patient as a random segmentation $L_{p,r}$ taking values in $\mathcal{S}$.
For each such random segmentation, we have access to $5$ observations $1\le i\le 5$, and denote each observation with
$l_{p,r}^{(i)}$.
The marginal function for patient $p$ and ROI $r$ is denoted by $m_{p,r}$ and formed by taking the point-wise average, that is
$m_{p,r}(\omega) = \frac{1}{5}\sum_{i=1}^5 l_{p,r}^{(i)}(\omega),\omega\in\Omega$.
In Figure~\ref{fig:ill}, two marginal functions formed this way are illustrated.
The experiment we run computes the following list of points:
\begin{align}
    \{ (\lVert s^{\mathrm{A}_{m_{p,r}}}\rVert_1/\lVert m_{p,r}\rVert_1, \lVert s^{\mathrm{D}_{m_p,r}}\rVert_1/\lVert m_{p,r} \rVert_1)\}_{1 \le p \le 19,1\le r \le 9}.
\end{align}
In Figure~5 theses points are illustrated as scatter plots and in Table~1 aggregated statistics of these points are listed.


\subsubsection{Comments on license, identifiability and consent}
From what we can gather, there is no licence specified by the creators.
However, it is specified in the article and in the download link (see below) that the data is free to use for non-commercial purposes.
As always is the case with medical image data, identifiable of the patients may be an issue.
Different experts and legislators have various opinions as to what lengths researchers has to go in order for the data to be considered anonymized.
However, in this case this is not an issue since all of the patients signed informed consent to be part of the data.

\subsubsection{Setup instructions}

\paragraph{1. Path: }
Before executing any code, make sure to set the file path to \path{Exierments_G}.
This folder will contain the code together with the data.

\paragraph{2. Data: }
The data used (Version 1, May 24 2017) can be acquired from the following link.
\begin{center}
\url{https://doi.org/10.5281/zenodo.583096}
\end{center}
Scroll down to the box labeled \emph{Files}, click on button labeled \emph{Request access...} and follow the instructions. 

Access will be granted provided your fulfill the criteria of requesting the data for academic or educational purposes.
When granted, an email will be sent to you with a link.
Follow this link and scroll down to the box labeled \emph{Files} again.
A list of files, each with a button labeled 
\emph{"Download"} should now be visible.
Download all of the files to the \path{Experiments_G/dicom} folder and rename
\path{3_03_P (1).zip}  to \path{3_03_P.zip}.
When done \path{Experiments_G/dicom}, should be populated with the following 19 files:
\path{1_01_P.zip},
\path{1_02_P.zip},
\path{1_03_P.zip},
\path{1_04_P.zip},
\path{1_05_P.zip},
\path{1_06_P.zip},
\path{1_07_P.zip},
\path{1_08_P.zip},
\path{2_03_P.zip},
\path{2_04_P.zip},
\path{2_05_P.zip},
\path{2_06_P.zip},
\path{2_09_P.zip},
\path{2_10_P.zip},
\path{2_11_P.zip},
\path{3_01_P.zip},
\path{3_02_P.zip},
\path{3_03_P.zip},
\path{3_04_P.zip}.

\paragraph{3. Plastimatch: }
For the computations, it is necessary to convert the segmentations from the \emph{rtstruct} format to the binary mask format \emph{nrrd}.
In this work, Plastimatch version 1.9.3 for Windows 64 which has a BSD-style license is used.
Both the installer and license can be found at the following address.
\begin{center}
    \url{http://plastimatch.org/}
\end{center}

For Ubuntu users, the Plastimatch software is available in the apt-repository.
There should be no reason as to why running it on this platform should be a problem, but it has not been tested.

\paragraph{4. Python: }
In this work, Python 3.10.4 is used.
It can be downloaded from the following address.
\begin{center}
    \url{https://www.python.org/downloads/}
\end{center}
Once in the right python environment, the necessary packages can be installed by using the provided requirements file.
\begin{verbatim}
pip install -r requirements.txt
\end{verbatim}

\paragraph{5. Running the code: }
In the \path{main.py} file, edit the variable \verb|plastimatch_match| so that it is compatible with the install path of Plastimatch.
The code is then simply executed with the following.
\begin{verbatim}
python main.py
\end{verbatim}
It will take approximately 30 min on a  descent desktop computer and should be no problems running on a laptop.
No GPU computations are done.
The code will start by unzipping all of the downloaded files to a temporary folder that will be deleted after the run.
It will then run Plastimatch to extract a mask for every available segmentation and put the results in \path{Experiments_G/masks}.
Once this is done, the discrete versions of the marginal functions are computed and used to compute the relative volumes.
When complete, the results of the experiment can be found under \path{Experiments_G/results}.

\subsection{Experiment L}

\subsubsection{Details on experiment}
\begin{figure}[htb!]
    \centering
\begin{tikzpicture}
  	\begin{axis}[
  	    name=plot11,height=6cm,width=6cm,
		xlabel=$\omega_1$, xtick={0,0.25,0.5,0.75,1.0},xmin=0,xmax=1,
		ylabel=$\omega_2$, ytick={0,0.25,0.5,0.75,1.0}, ymin=0,ymax=1, axis on top]
         \addplot graphics[xmin=0,xmax=1,ymin=0,ymax=1] {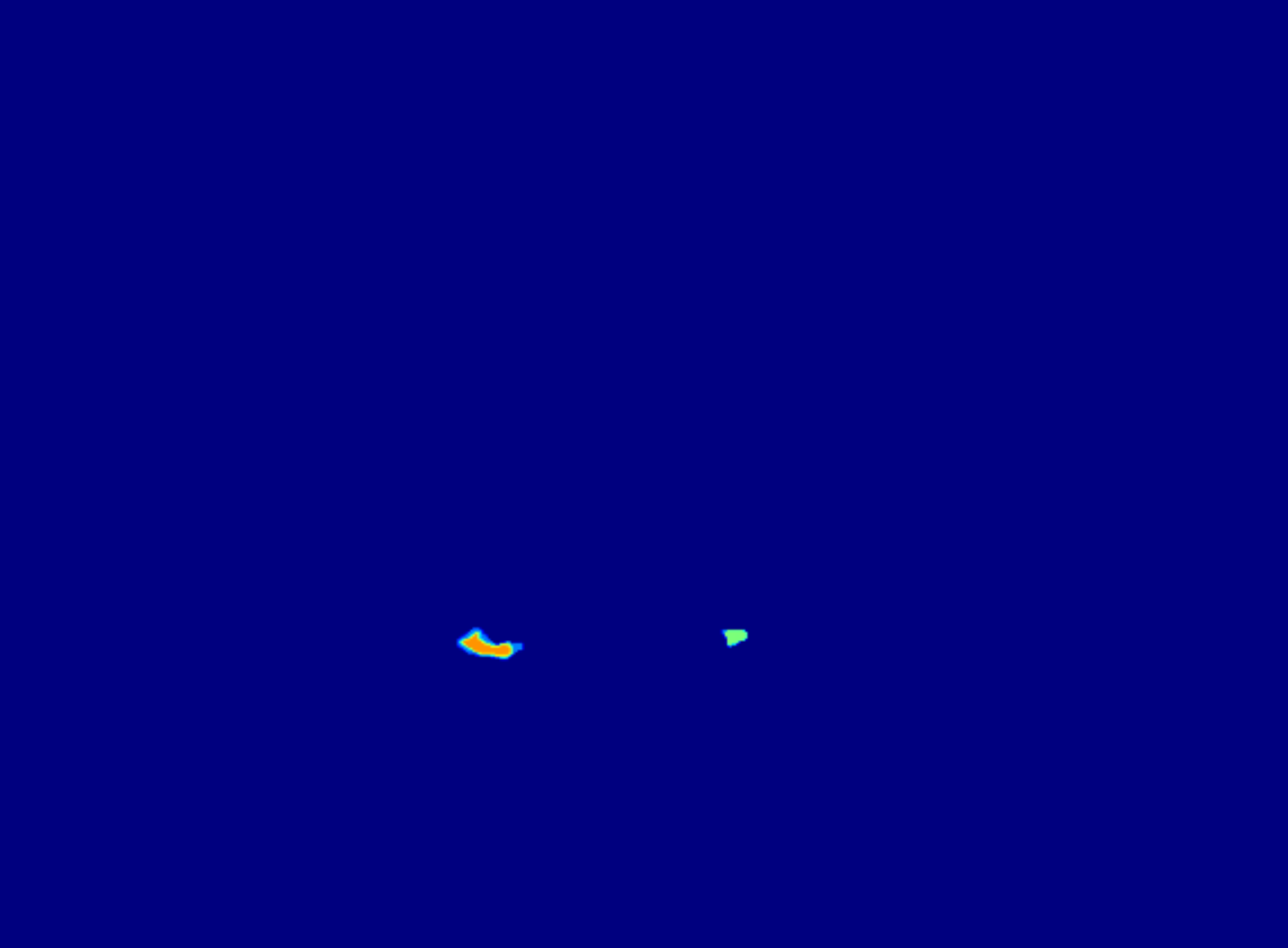};
	\end{axis}
  \begin{axis}[name=plot12,at={($(plot11.east)+(1.5cm,0)$)},anchor=west,height=6cm,width=6cm,
		xlabel=$\omega_1$, xtick={0,0.25,0.5,0.75,1.0},xmin=0,xmax=1 ,
		ylabel=$\omega_2$, ytick={0,0.25,0.5,0.75,1.0}, ymin=0,ymax=1, axis on top]
         \addplot graphics[xmin=0,xmax=1,ymin=0,ymax=1] {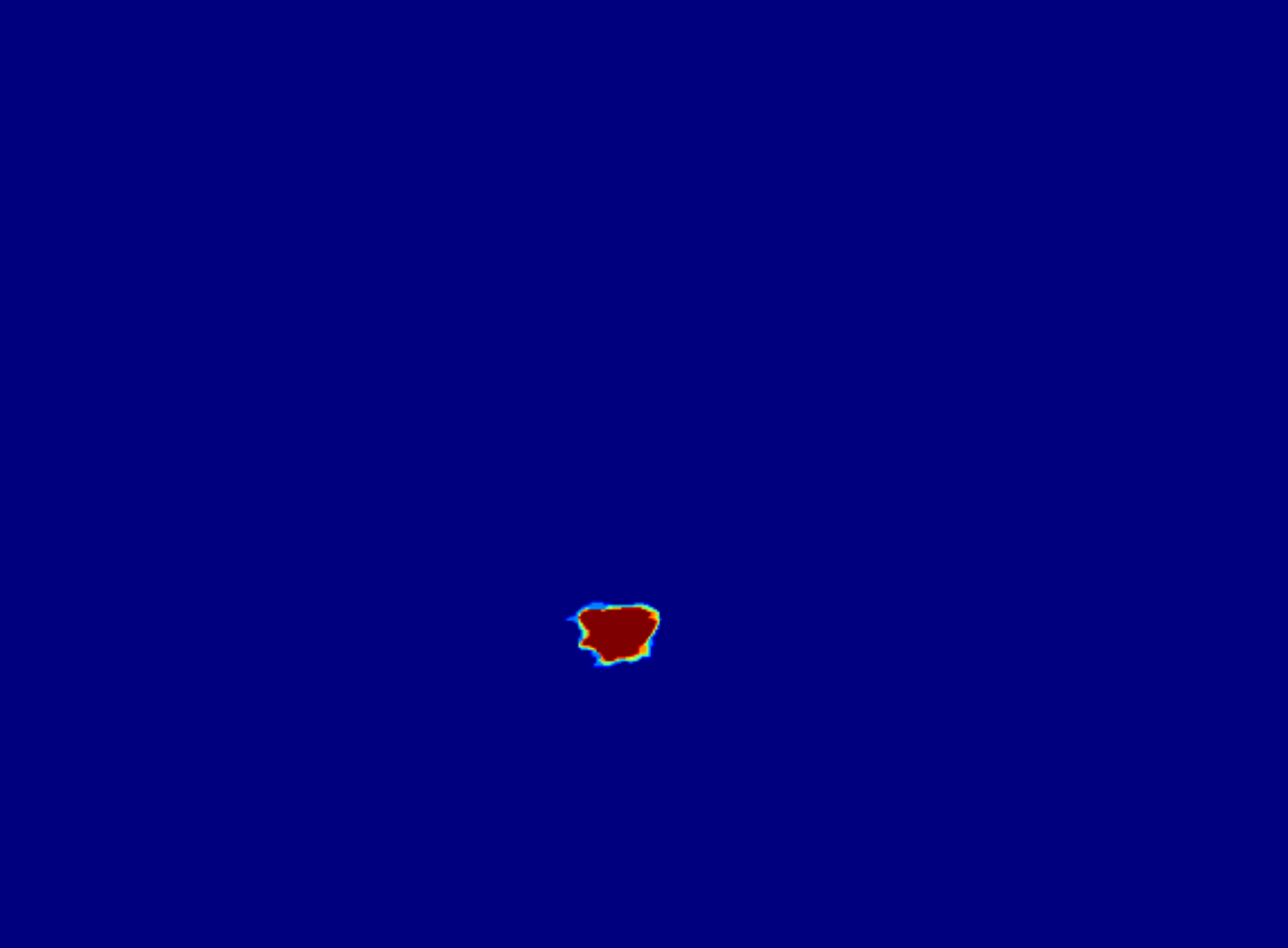};
  \end{axis}
  \begin{axis}[name=plot12,at={($(plot12.east)+(0.75cm,0)$)},anchor=west,height=6cm,width=2cm,
		xtick=\empty,
		ytick={0,0.25,0.5,0.75,1.0},
		yticklabel pos = right, xmin=0, xmax=1,ymin=0,ymax=1,axis on top]
 \addplot graphics[xmin=0,xmax=1,ymin=0,ymax=1] {raw/ill_colorbar_pgf.pdf};
  \end{axis}
    
\end{tikzpicture}
    \caption{Example of a slice from a marginal function where two lung nodules are visible.}
    \label{fig:ill_L}%
\end{figure}

The data consist of 1018 patients with lung nodules which have been delineated by 4 separate practitioners.
This leads to a total of 4072 segmentations.
Let $1 \le p \le 1018$ be an index of the patients.
We then think of the ROI for each patient as a random segmentation $L_{p}$ taking values in $\mathcal{S}$.
For each such random segmentation, we have access to $4$ observations $1\le i\le 4$, and denote each observation with
$l_{p}^{(i)}$.
The marginal function for patient $p$ is denoted by $m_{p}$ and formed by taking the point-wise average, that is
$m_{p}(\omega) = \frac{1}{4}\sum_{i=1}^4 l_{p}^{(i)}(\omega),\omega\in\Omega$.
In Figure~\ref{fig:ill_L}, a marginal functions formed this way is illustrated.
The experiment we run computes the following list of points:
\begin{align}
    \{ (\lVert s^{\mathrm{A}_{m_{p}}}\rVert_1/\lVert m_{p}\rVert_1, \lVert s^{\mathrm{D}_{m_p}}\rVert_1/\lVert m_{p} \rVert_1)\}_{1 \le p \le 1018}.
\end{align}
In Figure~6 these points are illustrated with scatter plots and in Table~1 aggregated statistics of these points are listed.

\subsubsection{Comments on license, identifiability and consent}
The data is licensed under Creative Commons Attribution 3.0 Unported License.
\begin{center}
\url{https://creativecommons.org/licenses/by/3.0/}.
\end{center}
Measures have been taken to anonymize the patient data.

\subsubsection{Setup instructions}
\paragraph{1. Path:}
Before executing any code, make sure to set the file path to \path{Experiments_L}.
This folder will contain the code together with the data when downloaded and processed.

\paragraph{2. Data:}
To get the data set, follow this link
\begin{center}
\url{https://doi.org/10.7937/K9/TCIA.2015.LO9QL9SX}
\end{center}
and download the \emph{Radiologist Annotations/Segmentations (XML)} file.
In our experiments, version 3 is used, which at the time of writing is the current version.
Extract the zip file to the \path{Experiments_L/xml} folder.
Note that the DICOM files are not necessary in order to run the experiments.

\paragraph{3. Python:}
In this work, Python 3.10.4 is used.
It can be downloaded from the following address.
\begin{center}
    \url{https://www.python.org/downloads/}
\end{center}
Once in the right python environment, the necessary packages can be installed by using the provided requirements file.
\begin{verbatim}
pip install -r requirements.txt
\end{verbatim}

\paragraph{4. Running the code:}
When the previous steps are completed, the code is executed by simply running the following.
\begin{verbatim}
python main.py
\end{verbatim}
It will take approximately 30min on a  descent desktop computer and should be no problems running on a laptop.
No GPU computations are done.
The code will use the pilidc library to query annotation data and generate masks.
The masks are used to compute the discrete versions of the marginal functions and used to compute the relative volumes.
When complete, the results of the experiment can be found under \path{Experiments_L/results}.